%% file: opd.tex
\documentclass[11pt, a4paper]{thuc3i}
\usepackage[sort&compress]{natbib}
\setheadertext{Rethinking On-Policy Distillation}

\usepackage[utf8]{inputenc} %
\usepackage[T1]{fontenc}    %
\usepackage{hyperref}       %
\usepackage{url}            %
\usepackage{tabularx}
\usepackage{makecell}
\usepackage{array,ragged2e,booktabs}
\newcolumntype{P}[1]{>{\RaggedRight\arraybackslash}p{#1}}
\usepackage{longtable}
\usepackage{amsfonts}       %
\usepackage{amsthm}         %

\usepackage{nicefrac}       %
\usepackage{microtype}      %
\usepackage{xcolor}         %

\usepackage{algorithm}
\usepackage{algorithmic}

\definecolor{darkblue}{rgb}{0, 0, 0.5}
\hypersetup{colorlinks=true, citecolor=darkblue, linkcolor=darkblue, urlcolor=darkblue}

\usepackage{xurl}

\usepackage{soul}
\usepackage{amsmath}
\usepackage{subcaption}
\usepackage{graphicx}
\usepackage{changes}
\usepackage{mathtools}
\usepackage{pifont}
\usepackage{tocloft}

\usepackage{colortbl}
\usepackage{wrapfig}
\usepackage{xspace}
\usepackage{multirow}
\usepackage{enumitem}

\usepackage{listings}
\usepackage{fontawesome5}

\usepackage{caption}

\usepackage{afterpage}
\usepackage{float}
\usepackage[edges]{forest}
\usepackage[normalem]{ulem}
\usepackage{bbding}
\usepackage[most,skins,theorems]{tcolorbox}
\usepackage{epigraph}

\usepackage[tikz]{bclogo}
\usepackage[framemethod=tikz]{mdframed}

\definecolor{lightorange}{HTML}{faa755}
\definecolor{lightblue}{RGB}{220,235,250}
\tcbset{
  takeawaysbox/.style={
    title=Takeaways,
    colback=lightblue!80,
    colframe=black,
    fonttitle=\bfseries\small,
    coltitle=white,
    colbacktitle=black,
    enhanced,
    attach boxed title to top left={xshift=2.5mm,yshift=-2.5mm},
    boxed title style={rounded corners, size=small, colframe=black, colback=black},
    width=\linewidth,
    arc=3.5mm
  }
}
\tcbset{
  taskbox/.style={
    title=Task Definition,
    colback=lightblue!80,
    colframe=black,
    fonttitle=\bfseries\small,
    coltitle=white,
    colbacktitle=black,
    enhanced,
    attach boxed title to top left={xshift=2.5mm,yshift=-2.5mm},
    boxed title style={rounded corners, size=small, colframe=black, colback=black},
    width=\linewidth,
    arc=2mm
  }
}
\definecolor{overviewbg}{HTML}{FFF3E0}
\definecolor{overviewframe}{HTML}{E65100}
\tcbset{
  overviewbox/.style={
    title=Overview,
    colback=overviewbg,
    colframe=overviewframe,
    fonttitle=\bfseries\small,
    coltitle=white,
    colbacktitle=overviewframe,
    enhanced,
    attach boxed title to top left={xshift=2.5mm,yshift=-2.5mm},
    boxed title style={rounded corners, size=small, colframe=overviewframe, colback=overviewframe},
    width=\linewidth,
    arc=3.5mm
  }
}

\definecolor{phenobg}{HTML}{EBF5FB}
\definecolor{phenoframe}{HTML}{2980B9}
\definecolor{mechbg}{HTML}{FEF5E7}
\definecolor{mechframe}{HTML}{E67E22}
\definecolor{recipebg}{HTML}{D5F5E3}
\definecolor{recipeframe}{HTML}{1E8449}
\definecolor{overviewdark}{HTML}{2C3E50}

\usepackage{cleveref}

\renewcommand{\today}{2026-04-15}

\usepackage[dvipsnames]{xcolor}

\title{Rethinking On-Policy Distillation of Large Language Models: Phenomenology, Mechanism, and Recipe}

\author{%
    Yaxuan Li$^{*1,2}$,
    Yuxin Zuo$^{*\dagger1}$,
    Bingxiang He$^{*\dagger1}$,
    Jinqian Zhang$^{1}$,
    Chaojun Xiao$^{\ddagger1}$,
    Cheng Qian$^{3}$,
    Tianyu Yu$^{1}$,
    Huan-ang Gao$^{1}$,
    Wenkai Yang$^{4}$,
    Zhiyuan Liu$^{\ddagger1}$,
    Ning Ding$^{\ddagger1}$\\
    $^{1}$Tsinghua University \quad
    $^{2}$ShanghaiTech University \quad
    $^{3}$University of Illinois Urbana-Champaign \quad
    $^{4}$Renmin University of China
    \vskip -1mm
    \textbf{$^*$Equal Contribution.} \quad 
    \textbf{$^\dagger$Project Lead.} \quad \textbf{$^\ddagger$Corresponding Authors.}
    \vskip -1mm
    \faGithub~{\fontfamily{ppl}\selectfont \textbf{Code:}}~\url{https://github.com/thunlp/OPD}. \\
    \vskip 1mm
    \faEnvelope[regular] \texttt{hebx24@mails.tsinghua.edu.cn, \{xcj,liuzy,dingning\}@tsinghua.edu.cn}
}

\input{Sections/0_Abstract}

\pretitlefigure{%
    \begin{center}
        \includegraphics[width=\linewidth]{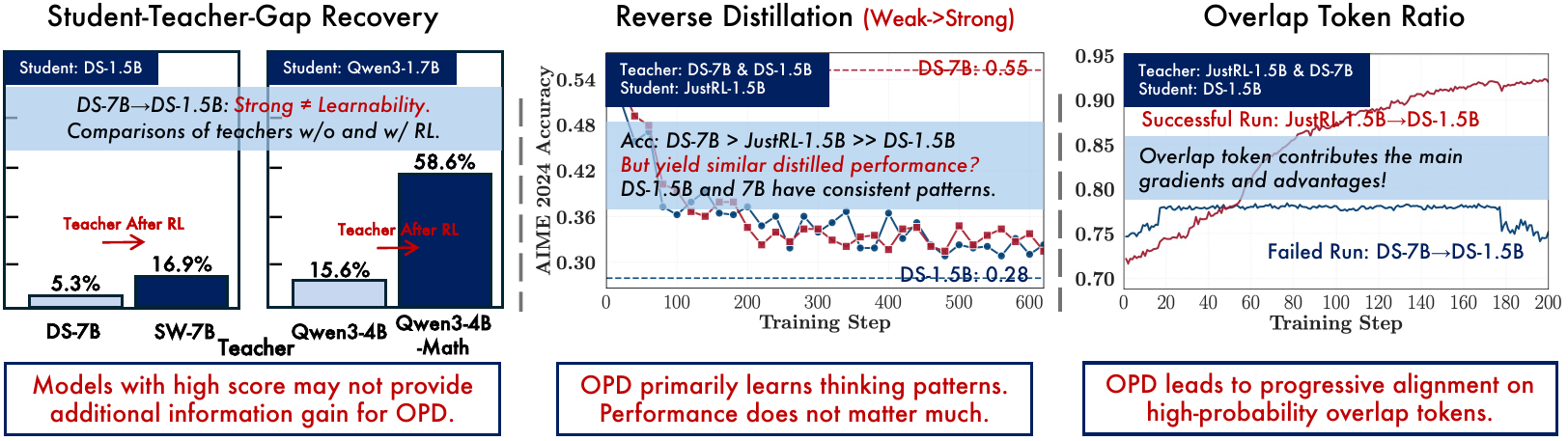}
        \captionof{figure}{Overview of our paper.
        JustRL-1.5B is obtained by applying RL to DeepSeek-Distill-1.5B~(DS-1.5B), and Skywork-OR1-Math-7B~(SW-7B) by applying RL to DeepSeek-Distill-7B~(DS-7B).
        }
        \label{fig:teaser}
    \end{center}
    \vskip4pt
}

\begin{document}

\maketitle

\input{Sections/1_Introduction}

\input{Sections/2_Preliminaries}

\input{Sections/3_Exp}
\input{Sections/4_Exp}

\input{Sections/5_Exp}
\input{Sections/6_Discussion}

\input{Sections/7_Related_Works}

\input{Sections/8_Conclusion}

\input{Sections/Others}
\bibliography{opd}

\newpage

\input{Sections/Appendix}

\end{document}

%% file: Sections/0_Abstract.tex
\begin{abstract}
On-policy distillation (OPD) has become a core technique in the post-training of large language models, yet its training dynamics remain poorly understood.
This paper provides a systematic investigation of OPD dynamics and mechanisms.
We first identify that two conditions govern whether OPD succeeds or fails: (i) the student and teacher should share compatible thinking patterns;
and (ii) even with consistent thinking patterns and higher scores, the teacher must offer genuinely new capabilities beyond what the student has seen during training.
We validate these findings through weak-to-strong reverse distillation, showing that same-family 1.5B and 7B teachers are distributionally indistinguishable from the student’s perspective.
Probing into the token-level mechanism, we show that successful OPD is characterized by progressive alignment on high-probability tokens at student-visited states, a small shared token set that concentrates most of the probability mass (97\%--99\%).
We further propose two practical strategies to recover failing OPD: off-policy cold start and teacher-aligned prompt selection.
Finally, we show that OPD's apparent free lunch of dense token-level reward comes at a cost, raising the question of whether OPD can scale to long-horizon distillation.

\end{abstract}

%% file: Sections/1_Introduction.tex
\section{Introduction}

On-policy distillation (OPD) has rapidly emerged as a core technique for large language model (LLM) post-training.
Recent industry efforts, including Qwen3~\citep{yang2025qwen3}, MiMo~\citep{xiao2026mimo} and GLM-5~\citep{zeng2026glm}, all adopt OPD in their post-training pipelines and report substantial gains, establishing it as a competitive complement to conventional supervised fine-tuning (SFT) and outcome-reward reinforcement learning (RL).
Thinking Machines Lab~\citep{lu2025onpolicydistillation} replicates the Qwen3 OPD recipe at a fraction of the RL compute cost, independently confirming that on-policy, dense supervision is a practically efficient alternative.

Unlike off-policy distillation, which trains the student on fixed teacher-generated sequences and suffers from exposure bias~\citep{bengio2015scheduled}, OPD has the student generate its own rollouts and leverages the teacher's per-token log-probabilities as a dense reward signal to refine behavior on states the student actually visits.
Recently, this has been extended to self-distillation settings where a single model serves as its own teacher given privileged information, demonstrating that the framework can drive continual self-improvement~\citep{hubotter2026reinforcement,zhao2026self,shenfeld2026self}.

However, despite these successes, OPD remains poorly understood and fragile in practice.
We observe a striking failure mode: a stronger teacher can completely fail to improve a student, even when a weaker teacher succeeds from lower initial alignment.
Yet few studies have investigated why the teacher's token-level signal steers the student distribution in the desired direction, or the conditions under which it fails.

\vspace{4pt}
\begin{tcolorbox}[
  enhanced,
  colback=white,
  colframe=black!25,          %
  boxrule=0.5pt,
  arc=2.5mm,                  %
  left=3mm, right=3mm,
  top=2.5mm, bottom=3mm,
]
\begin{center}
\small\textbf{\faBookOpen\hspace{0.4em}Paper Overview}
\end{center}
\vspace{-1mm}
\noindent
\begin{minipage}[c]{0.285\linewidth}
\parbox[t][1em][c]{\linewidth}{%
  \centering\scriptsize\color{gray}%
  \textit{When does OPD succeed or fail?}}
\begin{tcolorbox}[
  enhanced,
  colback=phenobg,
  colframe=phenoframe,
  fonttitle=\bfseries\footnotesize,
  title={\faChartLine\hspace{0.2em}~~Phenomenology (\S\ref{sec:phenomenology})},
  coltitle=white,
  colbacktitle=phenoframe,
  toptitle=1.3mm, bottomtitle=1.3mm,
  boxrule=0.4pt,
  arc=1.5mm,
  top=1.5mm, bottom=1.5mm, left=1.5mm, right=1.5mm,
  fontupper=\scriptsize,
  equal height group=overviewpillar,
]
Two empirical patterns distinguish effective OPD:\\[2pt]
{\scriptsize\color{black!70}%
\textbullet~Thinking-pattern consistency\\
\textbullet~Higher scores $\neq$ new knowledge}
\end{tcolorbox}
\end{minipage}%
\hfill
{\color{black!30}\raisebox{-2.5mm}{\large$\boldsymbol{\Rightarrow}$}}%
\hfill
\begin{minipage}[c]{0.285\linewidth}
\parbox[t][1em][c]{\linewidth}{%
  \centering\scriptsize\color{gray}%
  \textit{Why does OPD work at token level?}}
\begin{tcolorbox}[
  enhanced,
  colback=mechbg,
  colframe=mechframe,
  fonttitle=\bfseries\footnotesize,
  title={\faCogs\hspace{0.2em}~~Mechanism (\S\ref{sec:mechanism})},
  coltitle=white,
  colbacktitle=mechframe,
  toptitle=1.3mm, bottomtitle=1.3mm,
  boxrule=0.4pt,
  arc=1.5mm,
  top=1.5mm, bottom=1.5mm, left=1.5mm, right=1.5mm,
  fontupper=\scriptsize,
  equal height group=overviewpillar,
]
Progressive alignment on high-prob tokens governs OPD.\\[2pt]
{\scriptsize\color{black!70}%
\textbullet~Overlap tokens grow steadily\\
\textbullet~Overlap tokens alone suffice}
\end{tcolorbox}
\end{minipage}%
\hfill
{\color{black!30}\raisebox{-2.5mm}{\large$\boldsymbol{\Rightarrow}$}}%
\hfill
\begin{minipage}[c]{0.285\linewidth}
\parbox[t][1em][c]{\linewidth}{%
  \centering\scriptsize\color{gray}%
  \textit{How to rescue failing OPD?}}
\begin{tcolorbox}[
  enhanced,
  colback=recipebg,
  colframe=recipeframe,
  fonttitle=\bfseries\footnotesize,
  title={\faFlask\hspace{0.2em}~~Recipe (\S\ref{sec:recipe})},
  coltitle=white,
  colbacktitle=recipeframe,
  toptitle=1.3mm, bottomtitle=1.3mm,
  boxrule=0.4pt,
  arc=1.5mm,
  top=1.5mm, bottom=1.5mm, left=1.5mm, right=1.5mm,
  fontupper=\scriptsize,
  equal height group=overviewpillar,
]
Two strategies bridge the thinking-pattern gap:\\[2pt]
{\scriptsize\color{black!70}%
\textbullet~Off-policy cold start\\
\textbullet~Teacher-aligned prompts}
\end{tcolorbox}
\end{minipage}
\end{tcolorbox}
\vspace{2pt}

We present a systematic study of OPD training dynamics, progressing from empirical conditions through token-level mechanism to practical recipe.

\textbf{Phenomenology (\S\ref{sec:phenomenology}).}
We investigate the empirical patterns that distinguish effective from ineffective OPD, and identify two governing factors.
\emph{(i)~Thinking-pattern consistency}: the student and teacher should share consistent thinking patterns (e.g. higher overlap ratio in their top-$k$ token distributions). Even when the teacher achieves higher benchmark scores, mismatched thinking patterns produce low initial overlap that training cannot fully recover.
\emph{(ii)~Higher scores $\neq$ new knowledge}: even with consistent thinking patterns and higher benchmark scores, the teacher should offer knowledge that the student has not already acquired.
When both models are trained on the same data and recipe, they converge to similar distributions at their respective scales, leaving the teacher with little transferable signal.
Only when the teacher carries knowledge beyond what the student has already seen can OPD yield substantial gains.
We validate both conditions through reverse distillation experiments, which further reveal that OPD fundamentally learns thinking patterns rather than merely benefiting from pattern consistency, and that training dynamics can be entirely decoupled from benchmark scores.

\textbf{Mechanism (\S\ref{sec:mechanism}).}
We then investigate the token-level mechanism based on these conditions.
Across all settings studied, effective OPD exhibits a consistent signature where the student and teacher distributions become progressively more similar on student-visited states.
The high-probability tokens increasingly coincide (overlap ratio rising from 72\% to 91\%), the entropy gap between the two distributions narrows, and the shared top-$k$ tokens concentrate 97--99\% of the combined probability mass.
By contrast, failing runs show stagnant overlap and persistent entropy mismatch from the outset.
We further show that restricting supervision to overlap tokens alone matches full top-$k$ performance, confirming that the overlap set is the principal locus of OPD's gradient signal.

\textbf{Recipe (\S\ref{sec:recipe}).}
Guided by the phenomenological analysis, we propose two complementary strategies that recover OPD in otherwise failing configurations:
\emph{(i)~off-policy cold start}, a warmup SFT phase on teacher-generated rollouts before OPD that bridges the thinking-pattern gap by raising initial overlap ratio;
\emph{(ii)~teacher-aligned prompt selection}, which uses prompts drawn from the teacher's post-training data to sharpen alignment on high-probability tokens, though at the cost of substantially lower student entropy that necessitates mixing with out-of-distribution prompts.
In both cases, the recovered runs exhibit the same dynamic signature as naturally successful ones as shown in \S\ref{sec:mechanism}: steadily rising overlap ratio, improving token-level advantage, and a narrowing entropy gap.

Finally, we examine the cost of OPD's dense supervision (\S\ref{sec:discussion}).
We show that reward quality degrades systematically with trajectory depth and that instability originates at later tokens before propagating backward through the trajectory.
Surprisingly, even failing teachers provide reward that is globally correlated with rollout correctness, suggesting that the failure is not one of signal quality but of local optimization geometry. A larger teacher may induce a reward landscape that is locally flat around the student's policy, making token-level gradients ineffective despite an informative global signal.
These findings reveal a fundamental tension between supervision density and supervision reliability, and point to the limitations of current OPD for long-horizon reasoning and agentic settings.

%% file: Sections/2_Preliminaries.tex
\section{Preliminaries}

\subsection{Notation}

Let $x = (x_1, \ldots, x_n)$ denote an input prompt and
$y = (y_1, \ldots, y_m)$ a response.
We write $y_{<t} \triangleq (y_1, \ldots, y_{t-1})$ for the
prefix up to step $t$.
We consider two LLMs: a student $\pi_\theta$
and a teacher $\pi_T$, each defining a next-token distribution
$\pi(\cdot \mid x, y_{<t})$ over a vocabulary $\mathcal{V}$.
We write $y \sim \pi_\theta(\cdot \mid x)$ for a response
sampled autoregressively from the student.
$\mathcal{D} = \{(x^{(i)}, y^{(i)})\}_{i=1}^N$ denotes a
fixed dataset of prompt–response pairs with teacher-generated
outputs, and
$\mathcal{D}_x \triangleq \{x^{(i)}\}_{i=1}^N$ the
corresponding prompt set. Knowledge distillation (KD) transfers knowledge from $\pi_T$ to $\pi_\theta$ 
by minimizing the divergence between the two distributions. A standard choice is 
the Kullback-Leibler (KL) divergence, defined for two distributions $P$ and 
$Q$ over $\mathcal{V}$ as $D_{\mathrm{KL}}(P \| Q) 
  = \sum_{v \in \mathcal{V}} P(v) \log \frac{P(v)}{Q(v)}$.

\subsection{On-Policy Distillation}
\label{sec:on_policy_distillation}

On-Policy Distillation (OPD) computes supervision on trajectories
sampled from the current student $\pi_\theta$. Given a prompt
$x \sim \mathcal{D}_x$, the student samples a response
$\hat{y} = (\hat{y}_1, \ldots, \hat{y}_{T})
  \sim \pi_\theta(\cdot \mid x)$,
where $T \triangleq |\hat{y}|$ denotes the rollout length.
Both models are then evaluated on the student-generated prefixes
$\hat{y}_{<t}$, yielding two next-token distributions at each
step~$t$:
$p_t(v) \triangleq \pi_\theta(v \mid x, \hat{y}_{<t})$ and
$q_t(v) \triangleq \pi_T(v \mid x, \hat{y}_{<t})$
for $v \in \mathcal{V}$.

A standard formulation minimizes the sequence-level reverse KL over
student-generated trajectories:
\begin{equation}
  \mathcal{L}_{\mathrm{OPD}}(\theta)
  =
  \mathbb{E}_{x \sim \mathcal{D}_x}
  \Bigl[
    D_{\mathrm{KL}}\bigl(\pi_\theta(\cdot \mid x)\;\|\;\pi_T(\cdot \mid x)\bigr)
  \Bigr].
  \label{eq:opd_seq_rkl_prelim}
\end{equation}

Using the autoregressive factorization, this sequence-level objective admits the exact token-level decomposition:
\begin{equation}
  \mathcal{L}_{\mathrm{OPD}}(\theta)
  =
  \mathbb{E}_{x \sim \mathcal{D}_x,\;
    \hat{y} \sim \pi_\theta(\cdot \mid x)}
  \left[
    \sum_{t=1}^{T}
      D_{\mathrm{KL}}(p_t \| q_t)
  \right].
  \label{eq:opd_exact_token_decomp}
\end{equation}
In practice, different implementations vary
in how this exact per-token reverse KL is computed: full-vocabulary OPD optimizes Eq.~\eqref{eq:opd_exact_token_decomp} directly, sampled-token OPD uses an unbiased Monte Carlo estimator of each token-level KL term, and top-$k$ OPD replaces the full-vocabulary KL with a subset-based approximation.
We now describe these three common supervision granularities.

\paragraph{Sampled-Token OPD.}
The most lightweight variant evaluates only the token sampled by the student, and is also the most common implementation in prior on-policy distillation work~\citep{lu2025onpolicydistillation,xiao2026mimo,yang2026learning}. Given $\hat{y}_t \sim p_t$, the per-token loss is
$\ell_t^{\mathrm{sample}}
  \triangleq \log p_t(\hat{y}_t) - \log q_t(\hat{y}_t)$,
aggregated as:
\begin{equation}
  \mathcal{L}_{\mathrm{OPD}}^{\mathrm{sample}}(\theta)
  = \mathbb{E}_{x \sim \mathcal{D}_x,\;
      \hat{y} \sim \pi_\theta(\cdot \mid x)}
    \left[\sum_{t=1}^{T}
      \ell_t^{\mathrm{sample}}\right].
  \label{eq:opd_sampled_obj_prelim}
\end{equation}
Since
$\mathbb{E}_{\hat{y}_t \sim p_t}
  [\ell_t^{\mathrm{sample}}]
  = D_{\mathrm{KL}}(p_t \| q_t)$,
each $\ell_t^{\mathrm{sample}}$ is an unbiased single-sample
estimator of the token-level reverse KL.

\paragraph{Full-Vocabulary OPD.}
At the other extreme, one computes the divergence over the entire
vocabulary at each prefix:
\begin{equation}
  \mathcal{L}_{\mathrm{OPD}}^{\mathrm{full}}(\theta)
  = \mathbb{E}_{x \sim \mathcal{D}_x,\;
      \hat{y} \sim \pi_\theta(\cdot \mid x)}
    \left[\sum_{t=1}^{T}
      D_{\mathrm{KL}}(p_t \| q_t)\right].
  \label{eq:opd_full_obj_prelim}
\end{equation}
This yields denser gradients compared to sampled-token OPD, at the cost of $O(BTM)$ memory for batch size $B$, sequence length $T$ and vocabulary size $M = |\mathcal{V}|$.

\paragraph{Top-$k$ OPD.}
Top-$k$ OPD provides an intermediate design between sampled-token and
full-vocabulary OPD by restricting the divergence computation to a subset
$S_t \subseteq \mathcal{V}$. Here we focus on the student top-$k$ variant,
which selects the $k$ tokens assigned the highest probability under the
student, namely $S_t = \operatorname{TopK}(p_t, k)$.
Define the renormalized student and teacher distributions on $S_t$ as:
\[
\bar{p}_t^{(S_t)}(v)
=
\frac{p_t(v)\,\mathbf{1}[v \in S_t]}
     {\sum_{u \in S_t} p_t(u)},
\qquad
\bar{q}_t^{(S_t)}(v)
=
\frac{q_t(v)\,\mathbf{1}[v \in S_t]}
     {\sum_{u \in S_t} q_t(u)}.
\]
Distillation is then performed by minimizing the subset KL divergence $ D_{\mathrm{KL}}\!\bigl(\bar{p}_t^{(S_t)} \,\|\, \bar{q}_t^{(S_t)}\bigr)$, yielding the trajectory-level objective:
\begin{equation}
  \mathcal{L}_{\mathrm{OPD}}^{\mathrm{top\text{-}k}}(\theta)
  =
  \mathbb{E}_{x \sim \mathcal{D}_x,\;
      \hat{y} \sim \pi_\theta(\cdot \mid x)}
  \left[
    \sum_{t=1}^{T}
    D_{\mathrm{KL}}\!\bigl(\bar{p}_t^{(S_t)} \,\|\, \bar{q}_t^{(S_t)}\bigr)
  \right].
  \label{eq:opd_topk_obj_prelim}
\end{equation}

This formulation discards mass outside $S_t$ and therefore remains an approximation to the full-vocabulary reverse KL, but it substantially reduces teacher-query cost while preserving multi-token supervision on the student's high-probability region.

\subsection{Dynamic Metrics}
\label{sec:dynamic_metrics}

We define the student's and teacher's top-$k$ sets at step $t$
as $S_t^{(p)} = \operatorname{TopK}(p_t, k)$ and
$S_t^{(q)} = \operatorname{TopK}(q_t, k)$, respectively.
The following metrics are monitored throughout OPD training in later experiments.

\paragraph{Overlap Ratio.}
This metric quantifies the alignment between the student's and teacher's candidate spaces. It is defined as the average proportion of tokens that appear simultaneously in both the student's and the teacher's top-$k$ sets:
\begin{equation}
\mathcal{M}_{\text{overlap}}
\triangleq
\mathbb{E}_{t} \left[ \frac{|S_t^{(p)} \cap S_t^{(q)}|}{k} \right].
\label{eq:metric_overlap}
\end{equation}
A low overlap ratio indicates that the student's probability mass is concentrated on a disjoint set of tokens from the teacher, suggesting significant policy divergence or ``mode mismatch''. Conversely, a ratio nearing $1.0$ implies the student has successfully located the teacher's support region.

\paragraph{Overlap-Token Advantage.}
To measure distributional agreement within the overlap tokens,
we define
$A_t(v) \triangleq \bar{p}_t(v)
  (\log \bar{q}_t(v) - \log \bar{p}_t(v))$
where $\bar{p}_t, \bar{q}_t$ are the renormalized student and
teacher distributions over $S_t^{(p)} \cap S_t^{(q)}$.
The metric averages this quantity:
\begin{equation}
  \mathcal{M}_{\text{adv}}
  \triangleq \mathbb{E}_{t}\!\left[
      \frac{1}{|S_t^{(p)} \cap S_t^{(q)}|}
      \sum_{v \in S_t^{(p)} \cap S_t^{(q)}}
        A_t(v)
    \right].
  \label{eq:metric_adv}
\end{equation}
A value close to zero indicates high-quality alignment where the student places mass on teacher-preferred tokens with appropriate confidence. Conversely, a large negative value indicates that within the intersection, the student is overconfident compared to the teacher (high $p_t$ but lower $q_t$).

\paragraph{Entropy and Entropy Gap.}
To monitor the distributional properties of the policies, we track the entropy of both the student $H(p_t)$ and the teacher $H(q_t)$ on the student's rollouts, and define the entropy gap as:
\begin{equation}
\Delta H_t = \left| H(q_t) - H(p_t) \right|.
\label{eq:metric_entropy_gap}
\end{equation}
$\Delta H_t$ is a state-specific indicator of mode alignment.
A large gap suggests a substantial mismatch between the student and teacher in confidence and diversity over the same visited states, while convergence toward zero indicates that the student has matched the teacher's uncertainty profile along its generated trajectories.

%% file: Sections/3_Exp.tex
\section{Phenomenology of On-Policy Distillation}
\label{sec:phenomenology}

Before investigating the token-level mechanism of OPD, we first ask a broader question: what conditions govern the effectiveness of OPD?
A natural assumption is that a stronger teacher should always yield better distillation outcomes, yet we observe configurations where this fails.
We compare OPD runs under controlled settings and identify two conditions that jointly govern the outcome.

\begin{tcolorbox}[takeawaysbox]
\begin{itemize}[topsep=0pt, partopsep=0pt, leftmargin=12pt, itemsep=0pt]
    \item \textbf{Thinking-pattern consistency.} The student and teacher should share compatible thinking patterns. Even when the teacher achieves higher benchmark scores, a large mismatch weakens the token-level distillation signal (\Cref{sec:thinking_pattern_consistency}).
    \item \textbf{Higher scores $\neq$ new knowledge.} The teacher should provide knowledge beyond what the student has seen during training. Even with consistent thinking patterns and higher scores, a teacher may offer no genuinely new knowledge, leaving OPD without a driving signal (\Cref{sec:information_gain}).
\end{itemize}
\end{tcolorbox}

\subsection{Thinking-Pattern Consistency}
\label{sec:thinking_pattern_consistency}

\begin{figure*}[t]
    \centering
    \includegraphics[width=\textwidth]{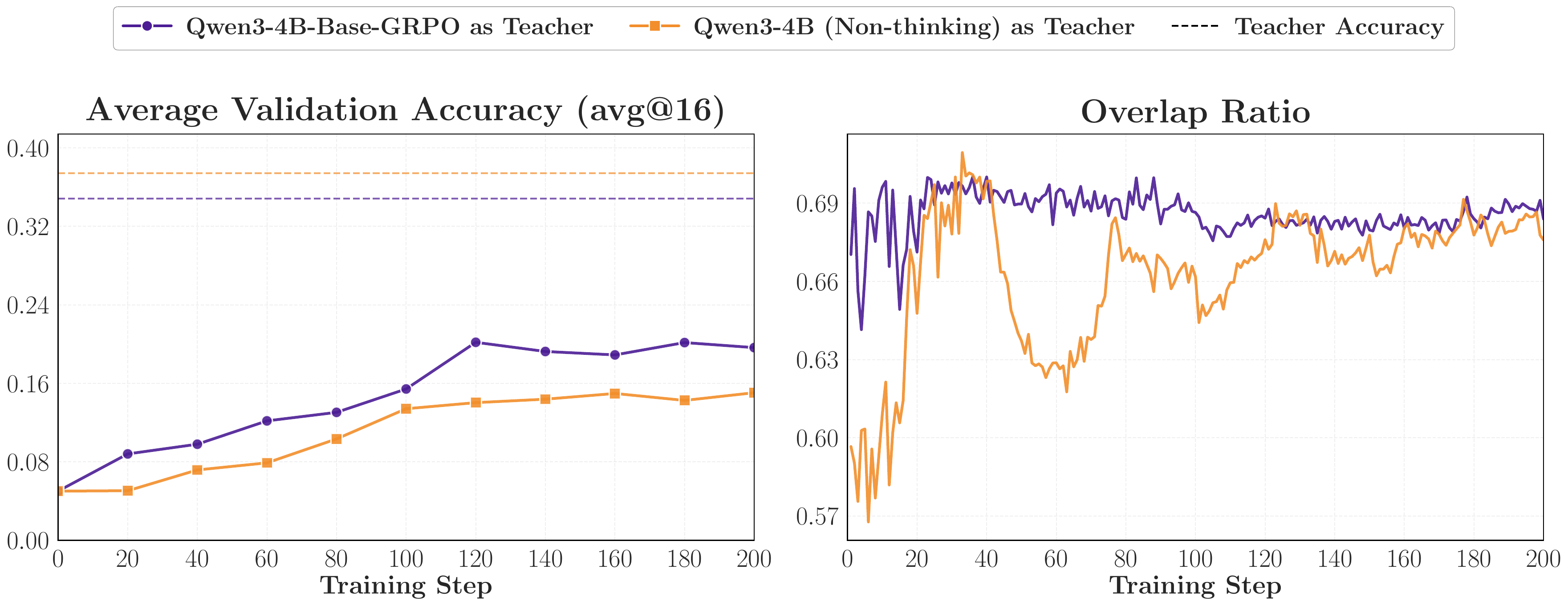}
    \caption{
    OPD from two teachers with different thinking patterns into the same student (Qwen3-1.7B-Base). The GRPO teacher achieves stronger performance (left) and higher initial overlap ratio (right), suggesting that thinking pattern compatibility governs OPD effectiveness.
    }
    \label{fig:thinking_pattern_compatibility}
\end{figure*}

We first study whether OPD requires compatible thinking patterns between the student and the teacher. A stronger teacher does not guarantee better distillation: a large mismatch in reasoning pattern can weaken the distillation signal regardless of the teacher's benchmark advantage.

\paragraph{Setup.}
We use Qwen3-1.7B-Base~\citep{yang2025qwen3} as the student and compare two teachers: Qwen3-4B (Non-thinking)~\citep{yang2025qwen3} and Qwen3-4B-Base-GRPO, where the latter is obtained by applying zero-RL to Qwen3-4B-Base~\citep{yang2025qwen3} using GRPO~\citep{shao2024deepseekmath} (detailed training settings are provided in Appendix~\ref{app:grpo_details}).
Since the student is also a base model, we expect its thinking pattern to be closer to that of the GRPO-trained teacher. We conduct two OPD experiments using the DAPO-Math-17K dataset~\citep{yu2025dapo}, differing only in the choice of teacher model. Unless otherwise specified, all experiments use the default hyperparameters described in Appendix~\ref{app:experimental_setup} and are evaluated on AIME 2024~\citep{li2024numinamath}, AIME 2025~\citep{balunovic2025matharena} and AMC 2023~\citep{li2024numinamath}.
Following standard practice, we sample 16 solutions per problem with temperature 0.7 and top-$p$ 0.95, using a maximum validation response length of 31{,}744 tokens. We report average accuracy over 16 samples (avg@16) as the primary evaluation metric.

\begin{wrapfigure}{r}{0.5\textwidth}
    \centering
    \includegraphics[width=\linewidth]{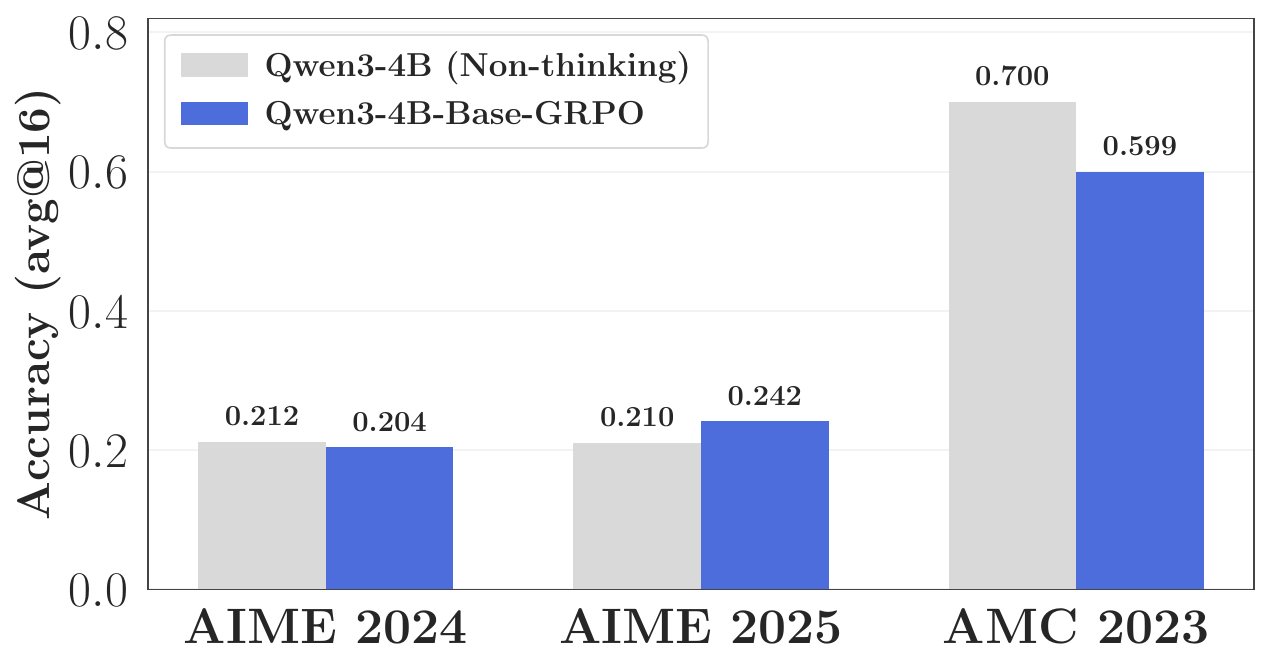}
    \caption{Validation performance of the two teachers (Qwen3-4B Non-thinking vs.\ Qwen3-4B-Base-GRPO) on AIME 2024, AIME 2025, and AMC 2023.
    }
    \label{fig:thinking_pattern_teacher_perf}
\end{wrapfigure}

\paragraph{Results.}
As shown in \Cref{fig:thinking_pattern_compatibility}, distillation from Qwen3-4B-Base-GRPO consistently outperforms distillation from Qwen3-4B (Non-thinking), although the two teachers have broadly comparable performance as shown in \Cref{fig:thinking_pattern_teacher_perf}. 
Despite underperforming on benchmarks, the GRPO teacher exhibits a higher initial overlap ratio, suggesting that its thinking pattern aligns more closely with the student.
Although the two overlap curves converge later in training, the performance gap persists, suggesting that early-stage thinking-pattern mismatch causes a loss of distillation benefit that cannot be recovered later.
We report the validation accuracy for each benchmark individually in Appendix~\ref{app:thinking_pattern_breakdown}, where the same overall trend holds across all datasets.

\subsection{New Knowledge, Not Just Scale}
\label{sec:information_gain}

Thinking-pattern consistency alone does not explain all of our observations. Even when the teacher scores higher and shares a consistent thinking pattern with the student, OPD can still fail.

\paragraph{Setup.}
We construct two controlled comparisons across model families.
In the DeepSeek family, we use DeepSeek-R1-Distill-Qwen-1.5B (R1-Distill-1.5B)~\citep{guo2025deepseek} as the student and compare two teachers: DeepSeek-R1-Distill-Qwen-7B (R1-Distill-7B)~\citep{guo2025deepseek} and Skywork-OR1-Math-7B~\citep{he2025skywork}, where the latter is obtained by applying RL post-training to R1-Distill-7B.
In the Qwen family, we use Qwen3-1.7B (Non-thinking)~\citep{yang2025qwen3} as the student and compare two teachers: Qwen3-4B (Non-thinking) and Qwen3-4B-Non-Thinking-RL-Math~\citep{yang2026learning}, where the latter is obtained by applying RL to Qwen3-4B (Non-thinking) on a 57K subset of DeepMath~\citep{he2025deepmath}.
In both settings, the key contrast lies between a teacher from the same training pipeline and one that has acquired additional capabilities through further RL. All runs use the same dataset and training recipe as before.

\begin{figure*}[t]
    \centering
    \begin{minipage}[t]{0.49\textwidth}
        \vspace{0pt}
        \centering
        \includegraphics[width=\linewidth]{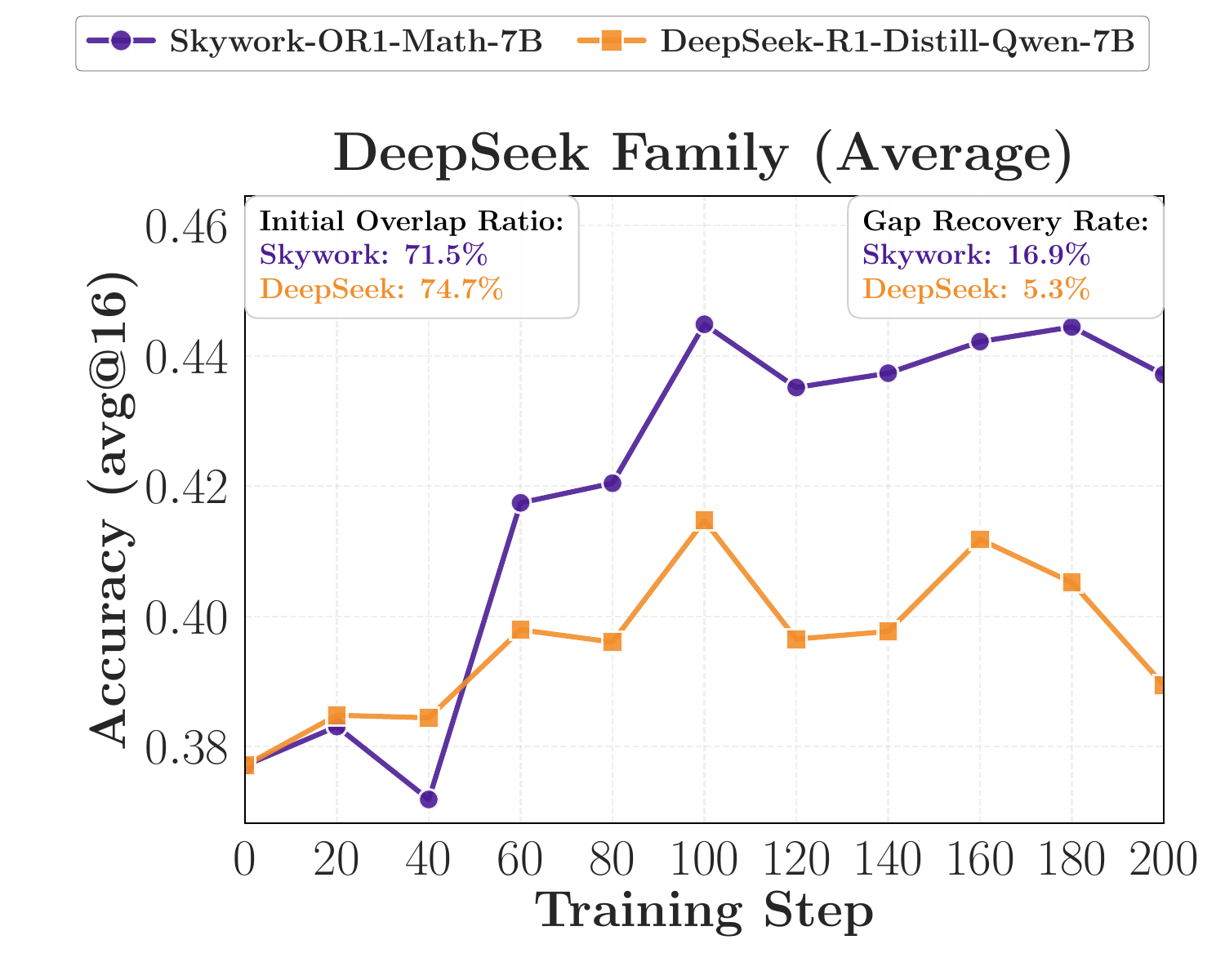}
    \end{minipage}
    \hfill
    \begin{minipage}[t]{0.49\textwidth}
        \vspace{0pt}
        \centering
        \includegraphics[width=\linewidth]{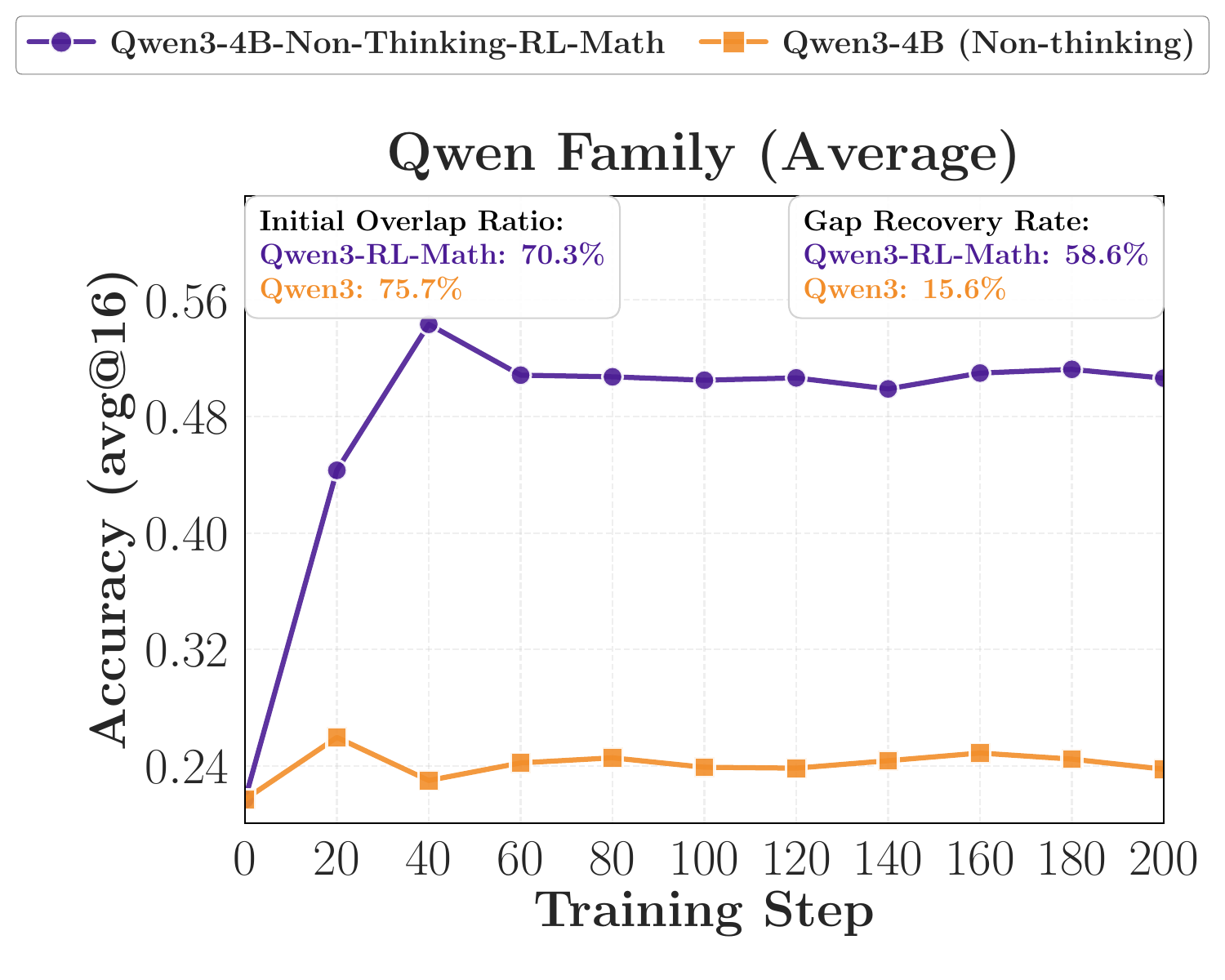}
    \end{minipage}
    \caption{
    Comparison of OPD performance with and without additional teacher RL post-training across two model families.
    \textbf{Left:} DeepSeek family. \textbf{Right:} Qwen family.
    Post-trained teachers yield substantially stronger gains and higher teacher-student \emph{gap recovery rate}.
    }
    \label{fig:information_gain_main}
\end{figure*}

\paragraph{Results.}
As shown in \Cref{fig:information_gain_main}, both families exhibit a consistent pattern. \textbf{\textcolor[HTML]{F3983E}{Same-pipeline teachers}} yield limited improvement, while the \textbf{\textcolor[HTML]{4E1F96}{post-trained teachers}} produce substantially stronger gains across all benchmarks.
Importantly, the post-trained teachers not only achieve higher absolute performance but also recover a much larger fraction of the teacher-student gap, measured by the \emph{gap recovery rate} $(\mathrm{Acc}_{\text{after OPD}} - \mathrm{Acc}_{\text{before OPD}}) / (\mathrm{Acc}_{\text{teacher}} - \mathrm{Acc}_{\text{before OPD}})$.
This indicates that the additional capabilities acquired by these teachers are more transferable through OPD.
Since the post-trained teachers are derived from the same base checkpoints, their thinking patterns remain broadly aligned, which is also observed by the overlap ratio dynamic.
The improvement therefore stems from new capabilities of the teacher acquired through RL.

\subsection{Validation via Reverse Distillation}
\label{sec:reverse_distillation}

We design a reverse-distillation experiment as the comparison that simultaneously validates both conditions and reveals deeper insights into the nature of OPD.

\paragraph{Setup.}
JustRL-DeepSeek-1.5B (JustRL-1.5B)~\citep{he2025justrl} is obtained by RL from R1-Distill-1.5B. We now reverse this direction, using JustRL-1.5B as the student and distilling from R1-Distill-1.5B (its own pre-RL checkpoint).
We also use R1-Distill-7B as a teacher for the comparison.
Note that R1-Distill-7B achieves slightly higher benchmark scores than JustRL-1.5B, while R1-Distill-1.5B is substantially weaker.

\begin{figure*}[t]
    \centering
    \includegraphics[width=\textwidth]{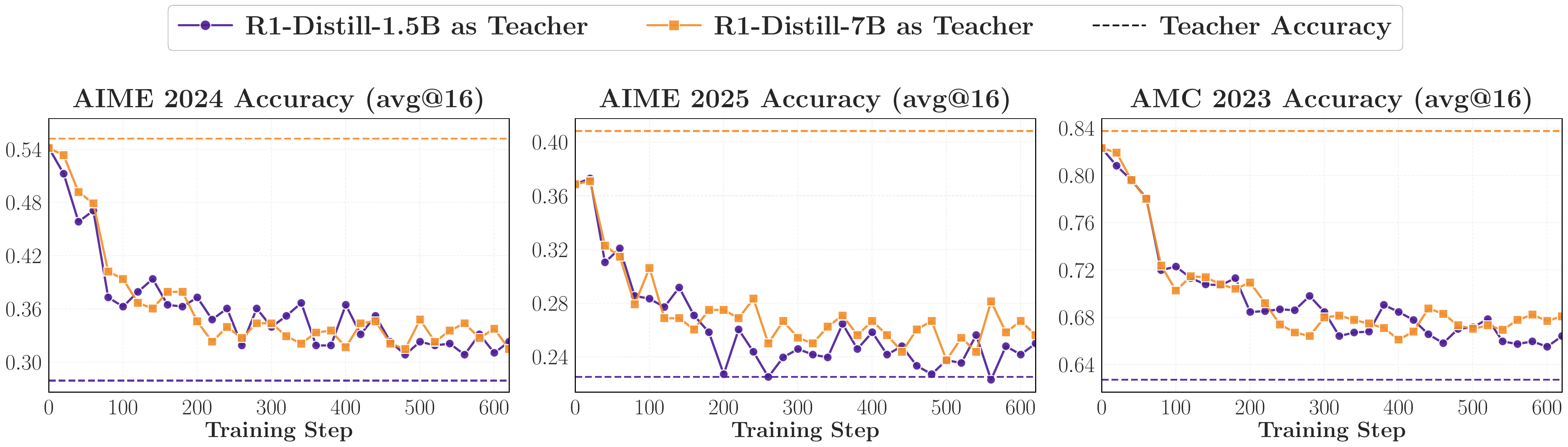}
    \caption{
    Reverse distillation with JustRL-1.5B as the student and two same-family teachers (R1-Distill-1.5B and R1-Distill-7B). Both runs cause the student to regress to approximately the same level despite R1-Distill-7B scoring higher than JustRL-1.5B, indicating that OPD training dynamics are governed by thinking pattern rather than benchmark performance.
    }
    \label{fig:information_gain_reverse}
\end{figure*}

\paragraph{Results.}
\Cref{fig:information_gain_reverse} reveals two striking phenomena.
First, distilling JustRL-1.5B toward R1-Distill-1.5B, its own pre-RL checkpoint, causes the student to regress almost exactly to its pre-RL performance, removing all gains acquired through RL.
Second, when we replace the teacher with R1-Distill-7B, a substantially larger and even slightly stronger model from the same family, the training trajectory is nearly indistinguishable: despite outscoring JustRL-1.5B on benchmarks, R1-Distill-7B drives the student to the same regressed level as the weaker 1.5B teacher.
Since OPD minimizes reverse KL divergence over student-generated trajectories, this convergence implies that the two teachers induce nearly identical local target distributions on student-visited states, despite their difference in scale.

These results yield several conclusions:
\begin{itemize}[leftmargin=12pt]
    \item \textbf{Thinking pattern matters, and OPD fundamentally learns thinking patterns.}
    Distilling from R1-Distill-1.5B into JustRL-1.5B causes JustRL-1.5B to regress to its pre-RL performance.
    This suggests that OPD actively acquires the teacher’s thinking patterns and overwrites the student's own. This is precisely why consistency in thinking patterns is important: if the gap is too large, the student may fail to learn effectively.
    
    \item \textbf{Benchmark performance does not predict OPD outcome.} R1-Distill-7B scores higher than JustRL-1.5B, yet the distillation produces no improvement and instead causes regression. This shows that OPD’s training dynamics can be completely independent of the teacher’s benchmark performance, and may even move in the opposite direction.
    
    \item \textbf{Higher scores do not imply new knowledge for OPD.} R1-Distill-7B and R1-Distill-1.5B are within the same model family and differ only in scale. The indistinguishable effects of the two models on the student already confirm that: (i) a higher score (R1-Distill-7B) may merely reflect a different degree of fit to the same data, rather than genuinely novel capabilities. For OPD to produce gains, the teacher should possess knowledge beyond what the student has already seen during training; and (ii) despite the difference in scale, R1-Distill-7B and 1.5B exhibit the same thinking patterns.

\end{itemize}

The reverse distillation experiments and the forward comparisons in \Cref{sec:thinking_pattern_consistency,sec:information_gain} consolidate the two conditions.
Thinking-pattern consistency is associated with higher initial overlap and stronger OPD outcomes, while new knowledge (such as from further post-training) enables larger transferable gains even when overlap is already high.

%% file: Sections/4_Exp.tex
\section{Mechanism of On-Policy Distillation}
\label{sec:mechanism}

\Cref{sec:phenomenology} identified two conditions, thinking-pattern consistency and new knowledge beyond the same model family, that govern OPD effectiveness.
We now investigate the token-level mechanism through which these conditions manifest during training.
By comparing successful and failing OPD runs, we show that effective distillation is driven by progressive alignment on high-probability tokens.

\begin{tcolorbox}[takeawaysbox]
\begin{itemize}[topsep=0pt, partopsep=0pt, leftmargin=12pt, itemsep=0pt]
    \item \textbf{Progressive alignment.} The overlap between the student’s and teacher’s high-probability top-$k$ tokens increases steadily throughout training at student-visited states; failing runs show stagnant overlap from the outset.
    \item \textbf{Overlap sufficiency.} Nearly all of the optimization's effect concentrates on the shared top-$k$ tokens; optimizing only these overlap tokens suffices to match standard OPD, while non-overlap tokens contribute little.
\end{itemize}
\end{tcolorbox}

\subsection{Progressive Alignment of High-Probability Tokens}
\label{sec:alignment_high_prob}

We compare the dynamics of a single student distilled from two different teachers under the same settings, one yielding clear improvement and the other yielding none.
We find that successful OPD is essentially driven by learning the high-probability tokens shared between the student and teacher.

\paragraph{Setup.} We choose R1-Distill-1.5B as the student and compare two teachers: JustRL-1.5B and R1-Distill-7B.
The two teachers exhibit comparable math performance, with the latter being slightly stronger. We use the same DAPO-Math-17K dataset and training settings as before, and monitor three dynamic metrics during OPD.

\begin{figure*}[t]
    \centering
    \includegraphics[width=\textwidth]{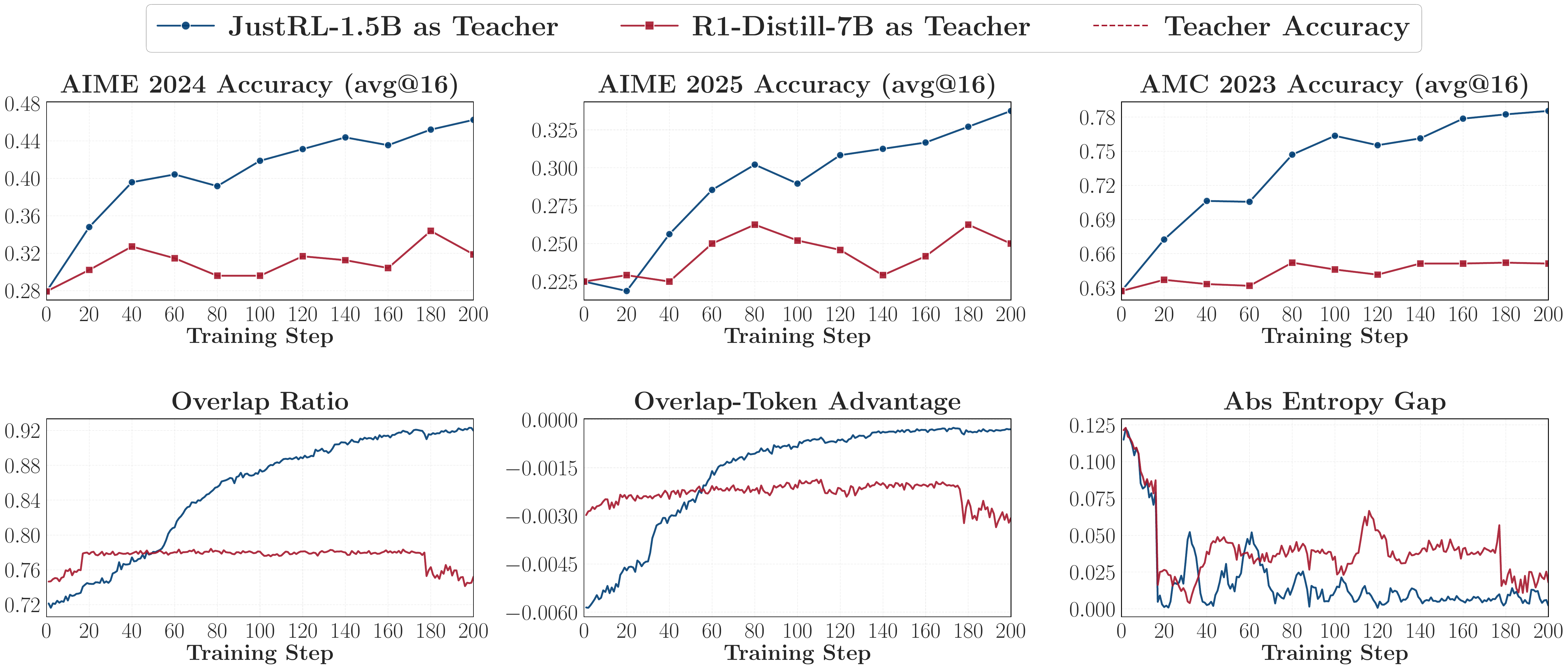}
    \caption{
    Successful vs.\ failing OPD with the same student (R1-Distill-1.5B) and two teachers.
    \textbf{Top:} avg@16 accuracy on three benchmarks. Dashed lines indicate teacher performance.
    \textbf{Bottom:} three dynamics over training. Successful distillation (JustRL-1.5B) shows rising overlap and narrowing entropy gap, and these trends are absent in the failing run (R1-Distill-7B).
    }
    \label{fig:alignment_main}
\end{figure*}

\paragraph{Results.}
\Cref{fig:alignment_main} shows sharply different outcomes.
Distillation from JustRL-1.5B yields consistent gains, with the final student recovering more than $80\%$ of the performance gap to the teacher, whereas distillation from R1-Distill-7B fails to yield any improvement despite the teacher being stronger overall.
The training dynamics (\Cref{fig:alignment_main}, bottom) reveal the underlying divergence.
\textbf{In the successful run, the overlap ratio rises steadily, the overlap-token advantage improves toward zero, and the entropy gap narrows, indicating that the student progressively locates the teacher's high-probability region, calibrates its mass within that region, and matches the teacher's local confidence. In the failing run, all three metrics stagnate.}

Two observations deserve emphasis. First, the overlap tokens carry $97\%$-$99\%$ of the total probability mass for both models throughout training (see Appendix~\ref{app:overlap_mass}), so the rising overlap reflects alignment on the probabilistically dominant tokens, not merely a set-level coincidence.
Second, the improvement in overlap-token advantage suggests that OPD's primary optimization signal lies in reweighting probability within the overlap region rather than in tokens outside it.

We also report auxiliary optimization metrics (policy loss, gradient norm, and extreme-advantage token probability differences) in Appendix~\ref{app:auxiliary_dynamics}, which show consistent secondary patterns: the successful run exhibits decreasing loss and sustained gradient magnitude, while the failing run shows weak gradients and persistent probability discrepancies.
We further verify that these findings generalize across different model pairs in Appendix~\ref{app:cross_model_validation}, using R1-Distill-7B as the student with two different teachers under the same settings.

\vspace{-10pt}
\subsection{Optimizing Shared Tokens Alone Suffices}
\label{sec:overlap_enough}

The above analysis shows that high-probability token alignment correlates with OPD success.
We further investigate whether this correlation is causal: whether the overlap region is not only where alignment emerges, but also the region that drives optimization.
We design a targeted ablation that decomposes the top-$k$ support into its overlap and non-overlap parts, training on each in isolation.

\textbf{Setup.} Using the successful OPD setting from \Cref{sec:alignment_high_prob} (JustRL-1.5B $\to$ R1-Distill-1.5B), we compare three variants that differ only in which tokens the distillation loss covers: (i) \textbf{Student Top-$k$}, which optimizes on the full student top-$k$ support $S_t^{(p)}$; (ii) \textbf{Overlap Top-$k$}, which restricts optimization to the intersection of the student and teacher top-$k$ sets $S_t^{(p)} \cap S_t^{(q)}$; and (iii) \textbf{Non-Overlap Top-$k$}, which restricts optimization to their symmetric difference $S_t^{(p)} \triangle S_t^{(q)}$. We set default $k$ to $16$.

\begin{figure*}[t]
    \centering
    \includegraphics[width=\textwidth]{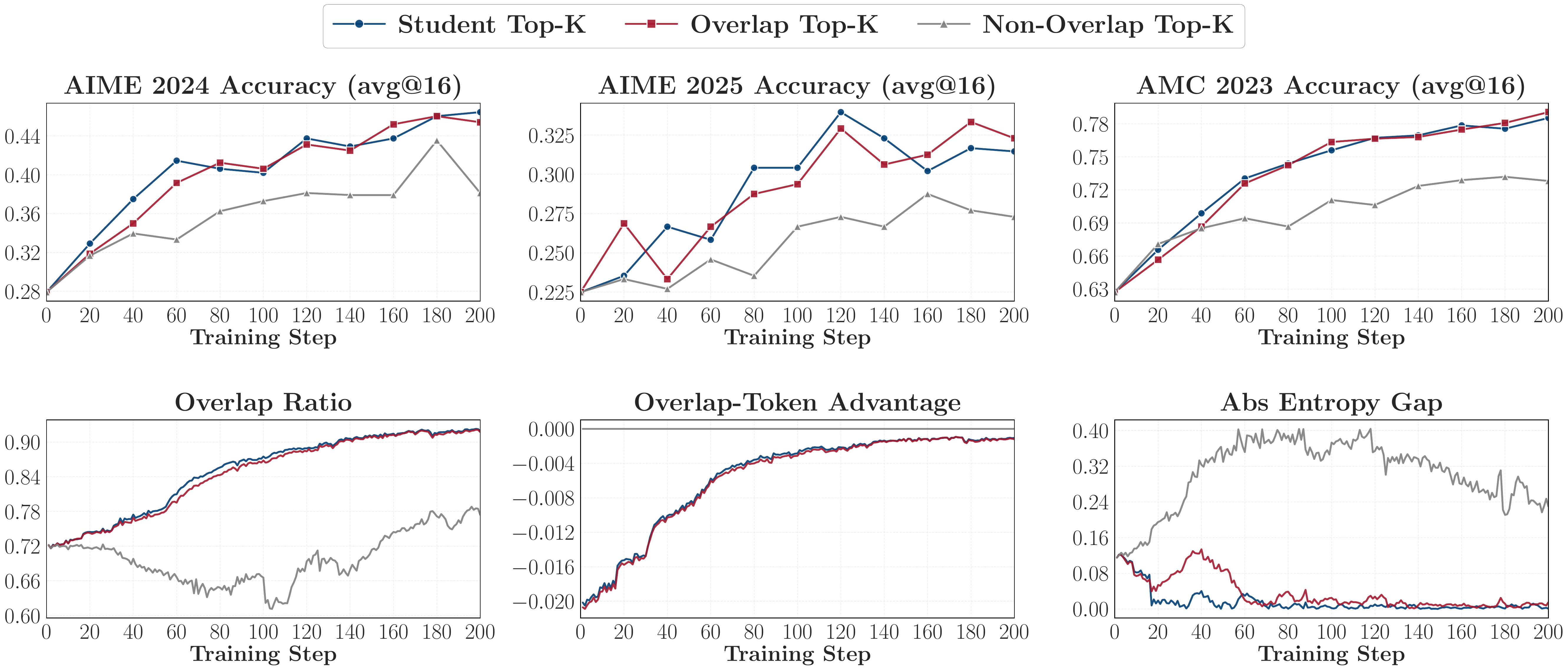}
    \caption{
    Ablation on the optimization support in Top-$k$ OPD. Overlap Top-$k$ matches Student Top-$k$, while Non-Overlap Top-$k$ is substantially weaker.
    }
    \vspace{-4mm}
    \label{fig:overlap_enough}
\end{figure*}

\paragraph{Results.} As shown in \Cref{fig:overlap_enough}, optimizing only the overlap region is sufficient to recover nearly the full benefit of standard Student Top-$k$ OPD on all three benchmarks, while Non-Overlap Top-$k$ remains consistently weaker.
This suggests that the main gains of OPD come from gradients on the shared high-probability region, rather than non-overlap tokens.
This also explains why Student Top-$k$ and Overlap Top-$k$ behave so similarly. The extra tokens in the student-only support carry very little probability mass. Consistently, the overlap-token advantage curves of Student Top-$k$ and Overlap Top-$k$ are nearly indistinguishable, whereas Non-Overlap Top-$k$ has much smaller magnitude, indicating a much weaker effective gradient on the overlap tokens.

\paragraph{Overlap optimization is self-reinforcing.}
Both Student Top-$k$ and Overlap Top-$k$ raise the overlap ratio steadily from about $72\%$ to above $91\%$, while Non-Overlap Top-$k$ first decreases and then only partially recovers (\Cref{fig:overlap_enough}, bottom-left).
This reveals a self-reinforcing dynamic: once a token enters the shared high-probability region and is favored by the teacher, reverse-KL updates concentrate more mass on it, gradually pushing competing non-overlap tokens out of the student's top-$k$ set. The overlap region thus grows not despite but because of the optimization, creating a virtuous cycle that sustains alignment throughout training.

Overall, these results support a unified mechanism for OPD: its primary effect is to progressively refine the student's distribution over teacher-supported high-probability tokens at student-visited states.
This alignment is both the signature of successful OPD and its operative locus, where optimizing only the overlap tokens suffices, and non-overlap tokens contribute little.
When the conditions identified in \Cref{sec:phenomenology} are met, this self-reinforcing dynamic drives steady improvement; when they are not, overlap stagnates and training fails to progress.

%% file: Sections/5_Exp.tex
\section{Practical Recipe}
\label{sec:recipe}

In \Cref{sec:phenomenology}, we identified two conditions for successful OPD.
While possessing new knowledge is an intrinsic property of the teacher, the thinking-pattern gap between the teacher and the student can be narrowed through training design.
In this section, we present two complementary strategies that recover OPD in otherwise failing configurations by improving the overlap dynamics.

\begin{tcolorbox}[takeawaysbox]
\begin{itemize}[topsep=0pt, partopsep=0pt, leftmargin=12pt, itemsep=0pt]
  \item \textbf{Off-policy cold start (\Cref{sec:off_distill_matters}).} Fine-tuning the student on teacher-generated rollouts before OPD closes the initial thinking-pattern gap, leading to higher overlap from the start and consistently stronger final performance.
  \item \textbf{Teacher-aligned prompts (\Cref{sec:teacher_posttrain_data}).} Using prompts from the teacher's post-training data sharpens alignment on high-probability tokens, although such prompts should be mixed with out-of-distribution prompts to prevent entropy collapse.
\end{itemize}

\end{tcolorbox}

\subsection{Off-Policy Distillation from Teacher Rollouts as Cold Start}
\label{sec:off_distill_matters}

When the student and teacher have substantially different thinking patterns, pure OPD can be ineffective because the teacher's token-level supervision is difficult for the student to exploit from its initial policy.
To mitigate this mismatch, we consider a two-stage framework: we first perform off-policy distillation by supervised fine-tuning (SFT) the student on teacher-generated rollouts to bring it closer to the teacher's thinking pattern, and then continue training with standard OPD.

\begin{figure*}[t]
    \centering
    \includegraphics[width=\textwidth]{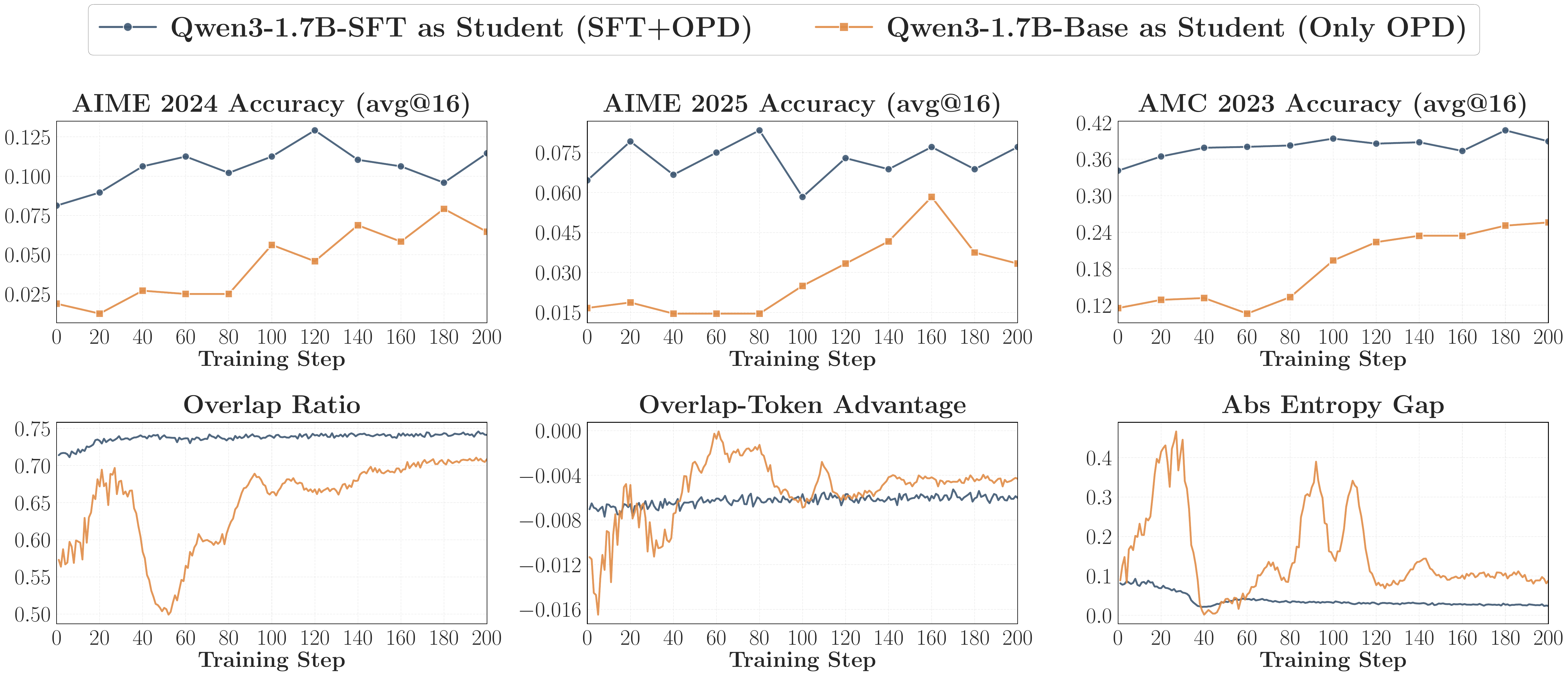}
    \caption{
    Effect of off-policy cold start before OPD, using fixed teacher Qwen3-4B (Non-thinking). The two curves correspond to OPD from Qwen3-1.7B-SFT and Qwen3-1.7B-Base.
    }
    \label{fig:off_policy_distillation}
\end{figure*}

\paragraph{Setup.}
We study this setting using Qwen3-1.7B-Base as the student and Qwen3-4B (Non-thinking) as the teacher.
We use the math-domain subset of OpenThoughts3-1.2M~\citep{guha2025openthoughts} as the prompt source for SFT.
The teacher generates 200K responses on a subset of this dataset, and we use these teacher rollouts to perform SFT on the student as a cold start, yielding Qwen3-1.7B-SFT.
We then continue training with OPD from this SFT initialization, using the remaining prompts from OpenThoughts after deduplicating against the SFT prompt subset (approximately 30K prompts).
As a control, we compare against a pure-OPD baseline that starts directly from Qwen3-1.7B-Base and uses the same teacher and OPD prompt set, but performs no cold-start distillation before OPD.
Detailed offline rollout and SFT configurations are provided in Appendix~\ref{app:coldstart_details}.

\paragraph{Results.}

As shown in \Cref{fig:off_policy_distillation}, the two-stage approach substantially outperforms pure OPD. Starting from Qwen3-1.7B-SFT yields consistently better validation performance than starting directly from Qwen3-1.7B-Base. Moreover, the performance gap persists throughout training, indicating that the off-policy cold start improves not only early optimization, but also the final performance ceiling of subsequent OPD.

The overlap dynamics support the same conclusion. The SFT-initialized student begins with a much higher overlap ratio and maintains a smooth, stable trajectory, whereas the base-initialized student starts lower and exhibits pronounced instability before gradually recovering.
The entropy gap is also substantially smaller for the SFT-initialized student, indicating a closer match to the teacher's confidence profile from the outset.
These observations confirm that off-policy distillation reduces the initial pattern mismatch, making the teacher's token-level supervision immediately exploitable once OPD begins.
A more detailed analysis of the overlap mass dynamics is provided in Appendix~\ref{app:add_analysis_of_overlap_mass}.

\subsection{Leveraging Teacher Post-Training Prompts}
\label{sec:teacher_posttrain_data}

The previous section narrows the thinking-pattern gap by moving the student closer to the teacher through SFT.
Another way to improve alignment is from the data side.
Since the teacher's policy is shaped by the prompts seen during post-training, we find that using teacher-aligned prompts during OPD yields more effective supervision.

\begin{figure*}[t]
    \centering
    \includegraphics[width=.95\textwidth]{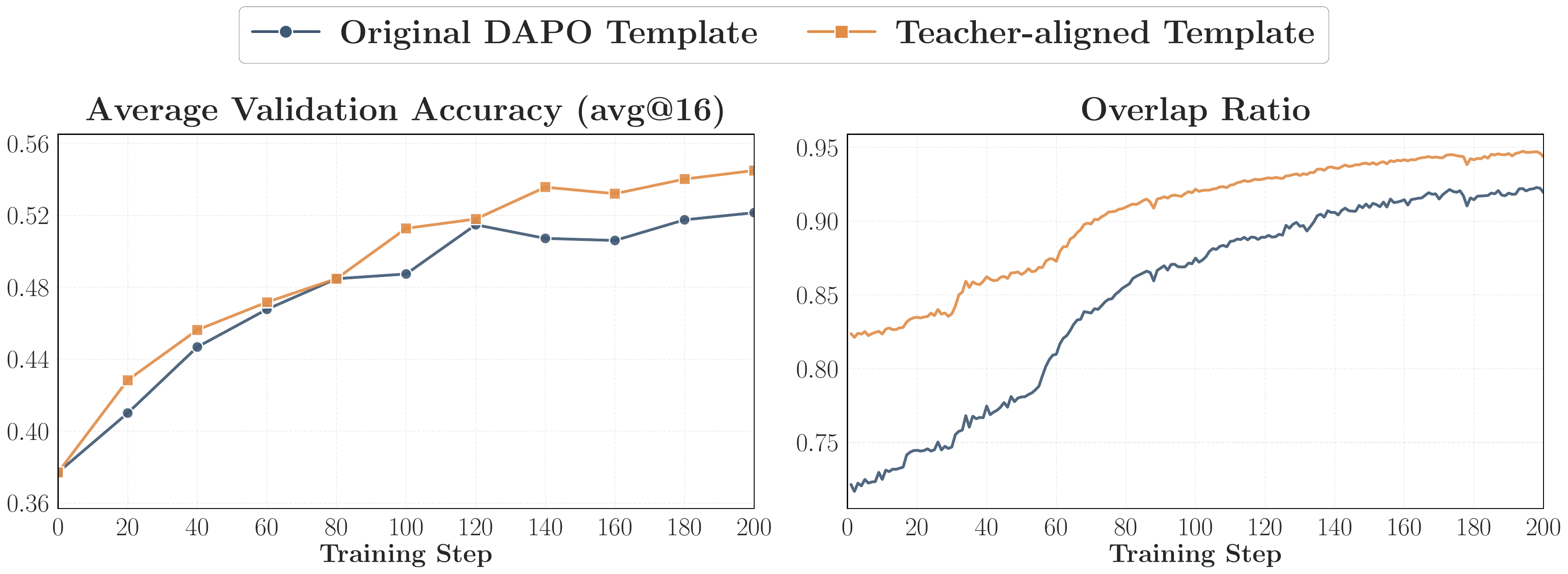}
    \caption{
    Effect of prompt template alignment. The teacher-aligned template yields higher accuracy and overlap growth throughout training.
    }
    \label{fig:teacher_posttrain_data_same}
\end{figure*}

\begin{figure*}[t]
    \centering
    \includegraphics[width=\textwidth]{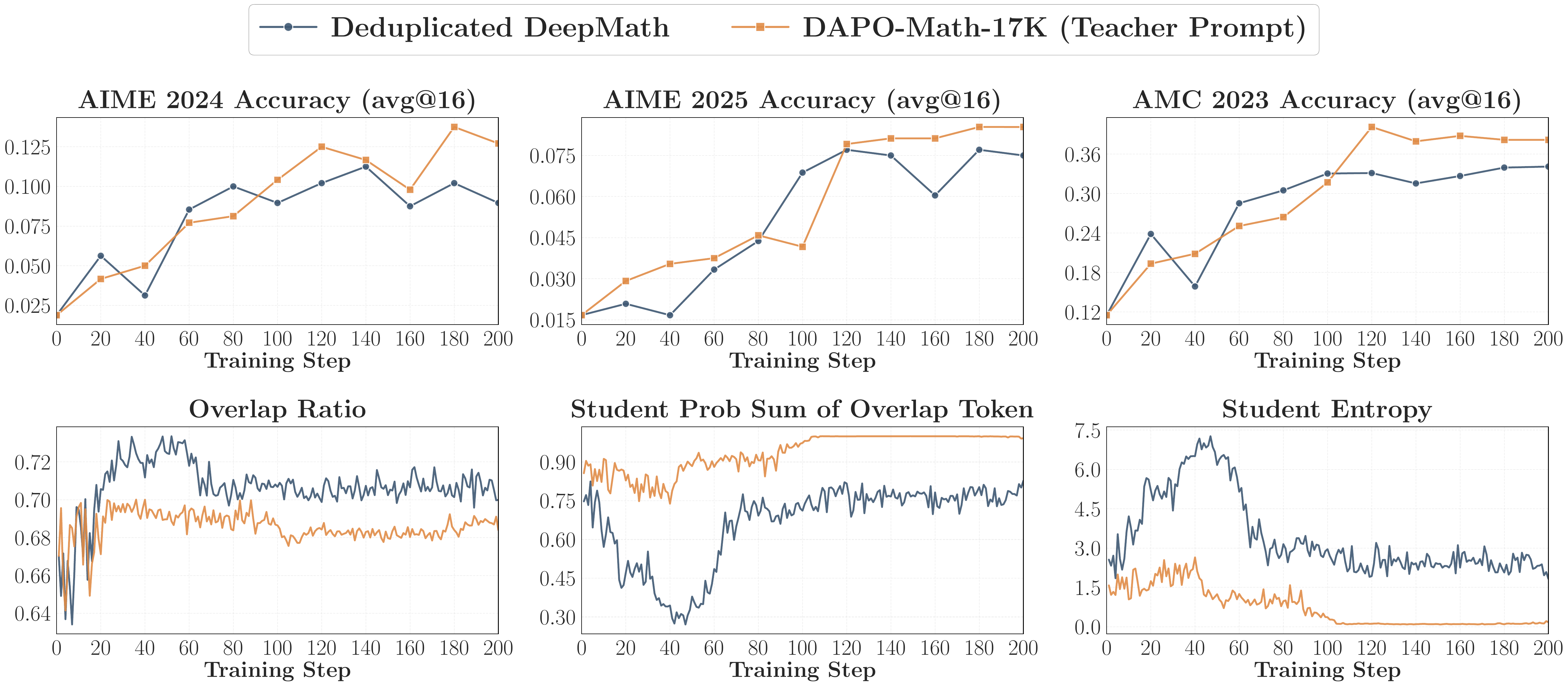}
    \caption{
    Effect of prompt content alignment. The teacher-aligned prompt contents yield stronger performance, with higher mass concentration on shared tokens and notably lower entropy.
    }
    \label{fig:teacher_posttrain_data_cross}
\end{figure*}

\paragraph{Setup.}

We conduct experiments at two granularities: whether matching the prompt \emph{template} matters, and whether matching the prompt \emph{content} matters.

\begin{itemize}[leftmargin=12pt]
    \item \textbf{Prompt template:} The teacher is JustRL-1.5B and the student is R1-Distill-1.5B. The prompt set is DAPO-Math-17K, with only the prompt template differing. The \emph{original} template is the standard DAPO format used in all previous experiments unless otherwise specified, while the \emph{teacher-aligned} template matches the format used during JustRL post-training:

\begin{tcolorbox}[title=\textbf{Original DAPO Template},colback=Salmon!20, colframe=Salmon!90!Black]
\small
Solve the following math problem step by step. The last line of your response should be of the form Answer: \$Answer (without quotes) where \$Answer is the answer to the problem. \{Question\} Remember to put your answer on its own line after ``Answer:''.
\end{tcolorbox}

\begin{tcolorbox}[title=\textbf{Teacher-Aligned Template}, colback=CornflowerBlue!20, colframe=CornflowerBlue!90!Black]
\small
\{Question\} Please reason step by step, and put your final answer within \textbackslash boxed\{\}.
\end{tcolorbox}

    Thus, the two runs contain the same math problems but differ in how the task is presented to the model. This design isolates the effect of prompt-template alignment with the teacher while keeping the underlying problem content unchanged.

    \item \textbf{Prompt content:} The teacher is Qwen3-4B-Base-GRPO introduced in \Cref{sec:thinking_pattern_consistency} and the student is Qwen3-1.7B-Base. We compare two prompt sets of matched size: DAPO-Math-17K (aligned with the teacher's RL training datasets) and a subset of DeepMath, deduplicated against DAPO-Math-17K (see Appendix~\ref{app:deepmath_dedup}). This design tests whether OPD benefits from using prompts that are identical to the teacher's post-training data, rather than prompts that are merely in-domain.
\end{itemize}

\paragraph{Results.}

The prompt template setting in \Cref{fig:teacher_posttrain_data_same} shows that simply switching to the teacher-aligned template improves validation performance on all three benchmarks.
The overlap dynamics support this result: the teacher-aligned template run begins with a higher overlap ratio and converges to a higher level, which indicates that even a minor change in prompt template can materially affect OPD by making the student's generated states more compatible with the teacher.
The benchmark-wise breakdown in Appendix~\ref{app:teacher_template_breakdown} shows the same trend.

The prompt content setting in \Cref{fig:teacher_posttrain_data_cross} shows a similar downstream advantage but with a subtlety: teacher-aligned prompts produce a lower overlap ratio throughout training. However, the cumulative student probability mass on the overlap tokens is substantially higher, indicating that the student concentrates its mass on fewer but more strongly shared tokens.
The effective alignment on high-probability tokens is therefore stronger, even though the overlap set is smaller.

\textbf{At the same time, we observe that using teacher-aligned prompts leads to substantially lower student entropy during training.}
This suggests that performing OPD only on prompts seen during teacher post-training may not always be ideal, as it can overly reduce policy entropy.
In practice, a more robust strategy may be to mix teacher-aligned prompts with prompts outside the teacher’s post-training data in order to preserve policy entropy and maintain the student’s capacity for exploration.

Overall, these results suggest that OPD benefits not only from an appropriate teacher, but also from a well-matched prompt set. Prompts closer to the teacher's post-training data can improve downstream performance and sharpen alignment on the most important shared tokens, but they should be used with care to avoid overly suppressing student entropy.

%% file: Sections/6_Discussion.tex
\section{Discussion}
\label{sec:discussion}

The appeal of OPD lies in its dense supervision, where every token receives a reward signal from the teacher, in contrast to the sparse outcome-level reward used in RL.
However, this increased supervision density comes at a cost.
The above sections all implicitly depend on the teacher's token-level reward being reliable in student-visited states, yet we have seen that this assumption can break down.
In this section, we investigate the reward signal itself and examine its properties and limitations.

\begin{figure}[t]
    \centering
    \includegraphics[width=\linewidth]{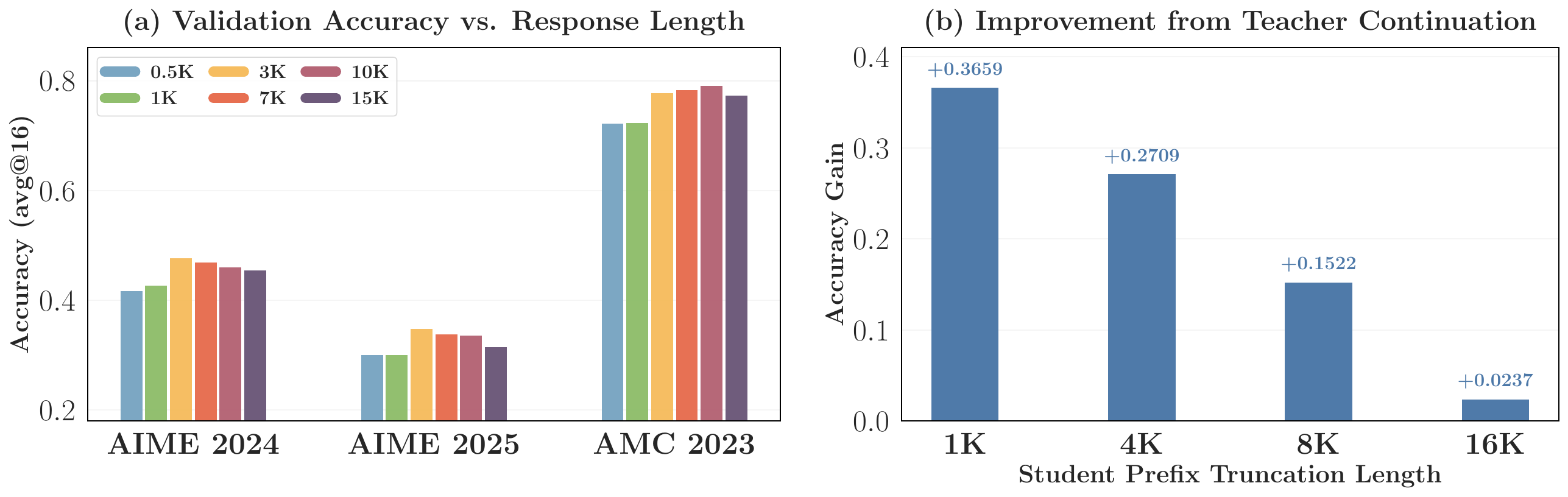}
    \caption{
    (a) Validation accuracy on three benchmarks under different response lengths.
    (b) Accuracy gain from teacher continuation under different student prefix truncation lengths.
    }
    \label{fig:response_length_combined}
\end{figure}

\begin{figure*}[t]
    \centering
    \includegraphics[width=\textwidth]{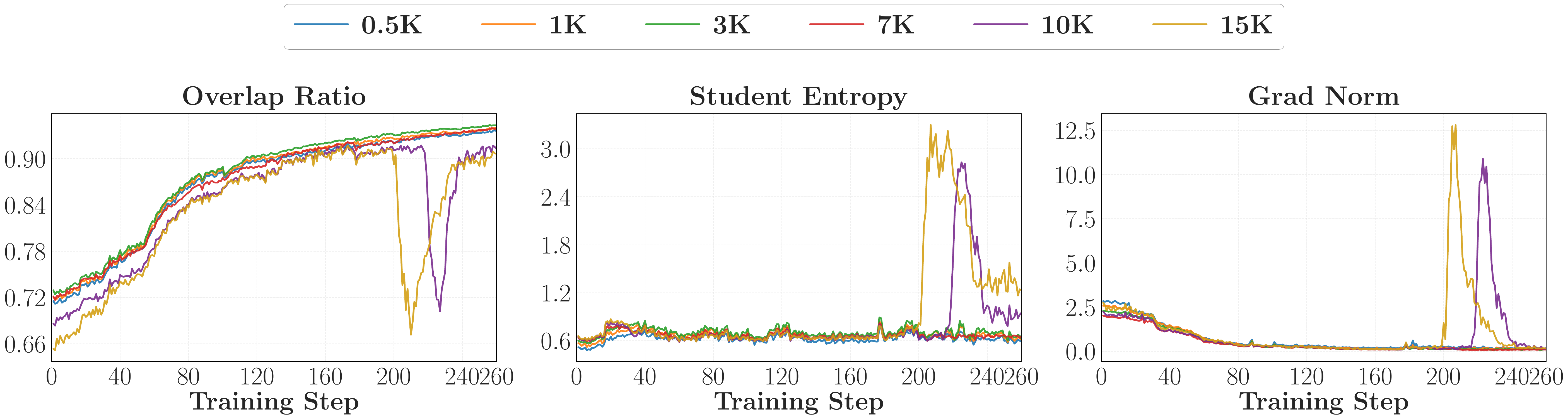}
    \caption{
    Training dynamics under different maximum response lengths for OPD.
    }
    \label{fig:response_length_dynamics}
\end{figure*}

\subsection{Reward Quality Degrades with Trajectory Depth}
\label{sec:reward_noise}

We first investigate how the teacher's reward quality varies with response length.

\paragraph{Response length exhibits a sweet spot.}

The supervision at position $t$ depends on the teacher's conditional $\pi_{T}(y_t \mid x, y_{<t})$ under a student-generated prefix $y_{<t}$, which may drift from trajectories the teacher would naturally produce. We train R1-Distill-1.5B against JustRL-1.5B across six maximum response lengths for 200 steps. As shown in \Cref{fig:response_length_combined}(a), very short responses (0.5K and 1K) provide too few supervised tokens for sample-efficient learning, while moderate lengths (3K and 7K) yield the strongest results. Beyond this range (10K and 15K), performance plateaus or declines. The training dynamics in \Cref{fig:response_length_dynamics} confirm that moderate lengths produce smooth overlap growth, whereas 10K and 15K exhibit late-stage collapse, with the overlap ratio dropping sharply, accompanied by spikes in student entropy and gradient norm.

\begin{figure*}[t]
    \centering
    \includegraphics[width=\textwidth]{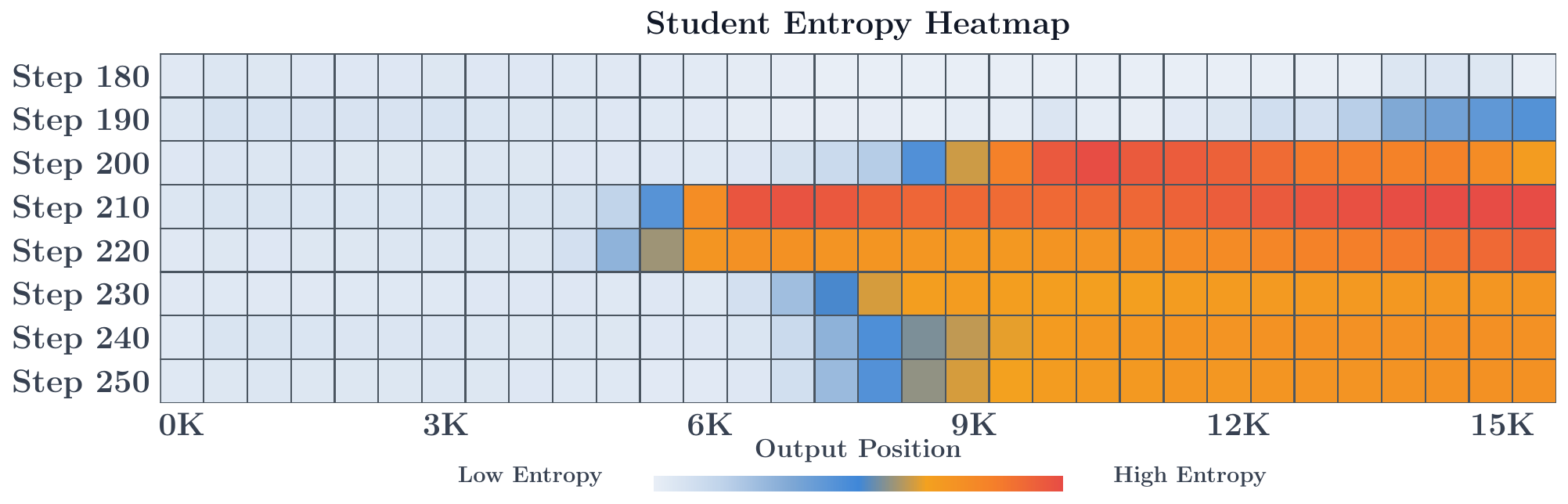}
    \caption{
    Average student entropy across decoding positions during OPD training with 15K max response length, measured on student-generated trajectories from Step 180 to Step 250.
    }
    \label{fig:student_entropy_by_position}
\end{figure*}

\paragraph{Instability originates at later tokens.}
Where does this collapse begin? In the 15K setting, analyzing student entropy as a function of output position across training steps reveals a clear back-to-front pattern: as shown in \Cref{fig:student_entropy_by_position}, high entropy first appears at the end of the response and progressively propagates toward earlier tokens as training proceeds. Teacher entropy exhibits a similar suffix-to-prefix trend (see Appendix~\ref{appendix:teacher_entropy_by_position}), consistent with the teacher encountering increasingly unfamiliar prefixes at later positions and producing noisier reward that in turn destabilizes the student.

\paragraph{Teacher continuation degrades with prefix depth.}
We further probe this by testing whether the teacher can still improve upon the student's continuation when starting from a student-generated prefix. We sample 2K prompts from DAPO-Math-17K, generate full student rollouts, and select those exceeding 16K tokens. We then truncate each rollout at multiple positions and let the teacher continue from the resulting prefix. \Cref{fig:response_length_combined}(b) shows that the teacher’s accuracy advantage decreases monotonically, from +0.37 at a 1K prefix to just +0.02 at a 16K prefix.

Together, these results reveal a fundamental tradeoff in OPD's token-level supervision. Dense reward is effective on moderately long reasoning traces, but its reliability degrades with depth as the student prefix drifts further from the states familiar to the teacher. This suggests that OPD may not extend cleanly to longer-horizon settings such as extended chain-of-thought or agentic multi-turn interaction.

\begin{figure*}[t]
    \centering
    \includegraphics[width=.9\textwidth]{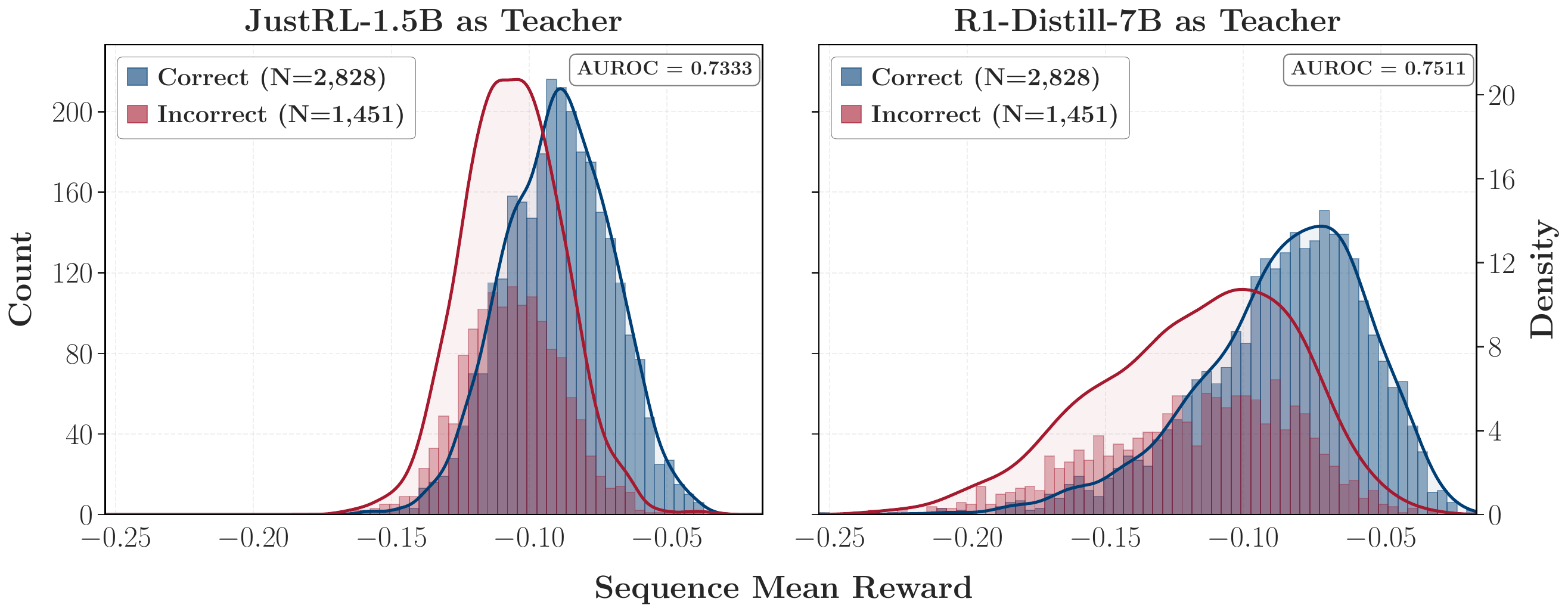}
    \caption{
    Sequence mean reward distributions for correct and incorrect student rollouts. Both teachers assign higher reward to correct rollouts with comparable AUROC (0.73 and 0.75).
    }
    \label{fig:reward_landscape}
\end{figure*}

\subsection{Globally Informative Reward Does Not Guarantee Local Exploitability}
\label{sec:landscape}

The previous subsection shows that reward quality degrades with trajectory depth. A natural follow-up question is whether the reward signal is fundamentally uninformative in failing OPD configurations or whether the source of failure lies elsewhere.

\paragraph{Setup.}
We revisit the controlled comparison from \Cref{sec:alignment_high_prob}, with R1-Distill-1.5B as the student and two teachers: JustRL-1.5B (successful OPD) and R1-Distill-7B (failed OPD). For each student rollout $y$, we compute the sequence mean reward
$\bar{r}(y) = \frac{1}{T}\sum_{t=1}^{T} \left[\log \pi_{T}(y_t \mid x, y_{<t}) - \log \pi_\theta(y_t \mid x, y_{<t})\right]$
under sampled-token OPD, and compare the distribution of $\bar{r}(y)$ between correct and incorrect rollouts.

\paragraph{Global reward structure is preserved in both settings.}
\Cref{fig:reward_landscape} shows that, for both teachers, correct rollouts consistently receive higher sequence mean reward than incorrect ones, with comparable AUROC values (0.73 for JustRL-1.5B, 0.75 for R1-Distill-7B). The failing 7B teacher does not produce a weaker global signal, which is equally correlated with rollout correctness.

\paragraph{A hypothesis on local optimization geometry.}
If the reward is globally informative in both cases, why does OPD fail with the 7B teacher? The training dynamics from \Cref{sec:alignment_high_prob} offer a clue. As shown in \Cref{fig:alignment_main}, when R1-Distill-7B serves as the teacher, the overlap-token advantage becomes larger in magnitude than in the JustRL setting during the later stages of training, yet the gradient norm remains persistently smaller (see Appendix~\ref{app:auxiliary_dynamics}). One possible explanation is that the 7B teacher's per-token advantages, while individually large, are anisotropic across positions within each sequence. When these heterogeneous signals are aggregated into a gradient update, they partially cancel, yielding small effective gradients despite large per-token rewards. By contrast, JustRL-1.5B, which shares a compatible thinking pattern with the student, may concentrate its advantage on a more coherent subset of tokens. The resulting gradient, though composed of smaller per-token signals, points in a consistent direction that reverse KL can amplify through its mode-seeking behavior.

We have not directly verified this anisotropy hypothesis, and doing so would require analyzing the directional structure of per-token gradients, which we leave to future work. Nonetheless, the co-occurrence of high per-token advantage and low gradient norm is suggestive and points to an important distinction that a globally informative reward does not guarantee a locally exploitable one. Understanding the geometry of OPD's reward landscape, and developing objectives that can exploit anisotropic reward structures, remains an open question.

\subsection{Sampled-Token Reward Is Already Sufficient}
\label{sec:optimal_top_k}

A natural question about OPD's reward is how many tokens per position are needed to compute a useful gradient. Top-$k$ OPD aggregates the reward over the $k$ highest-probability tokens at each position, and one might expect that larger support always leads to better or more stable learning. We investigate this by varying $k$ and comparing against the simpler sampled-token OPD, which uses only a single token drawn from the student distribution at each position.

\begin{figure*}[t]
    \centering
    \includegraphics[width=.75\textwidth]{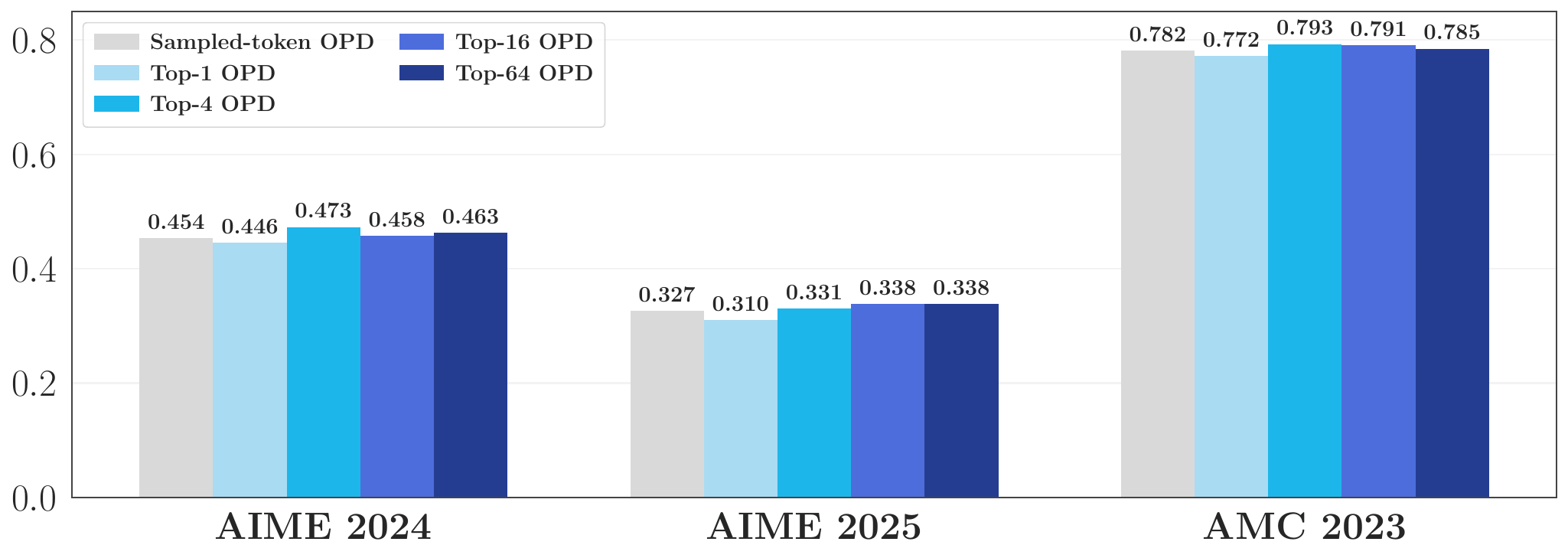}
    \caption{Effect of the support size $k$ in Top-$k$ OPD. All numbers are reported as avg@16.}
    \label{fig:effect_of_k_topk}
\end{figure*}

\paragraph{Setup.}
We use R1-Distill-1.5B as the student and JustRL-1.5B as the teacher, and compare Top-$k$ OPD with $k\in\{1, 4, 16, 64\}$ against sampled-token OPD, keeping all other hyperparameters fixed.

\begin{figure*}[t]
    \centering
    \includegraphics[width=\textwidth]{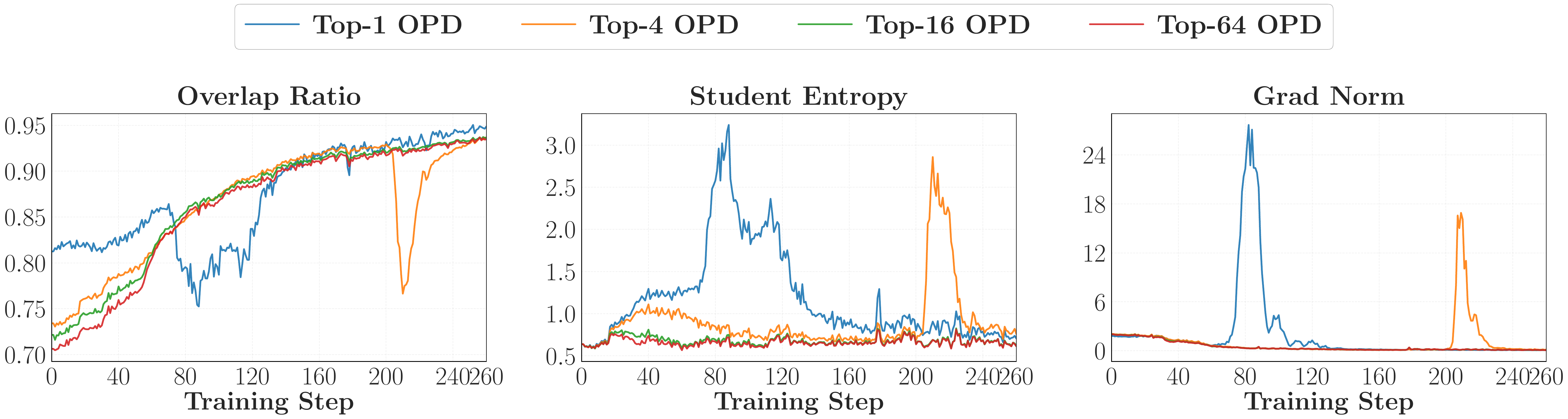}
    \caption{Training dynamics under different support sizes $k$ for Top-$k$ OPD.
    }
    \label{fig:effect_of_k_topk_dynamics}
\end{figure*}

\paragraph{Results.}
\Cref{fig:effect_of_k_topk} shows that sampled-token OPD achieves performance comparable to that of the Top-$k$ settings averaged on three benchmarks. The only clearly worse configuration is Top-$1$, which consistently underperforms. Enlarging $k$ beyond 4 brings negligible additional gain while leading to greater computational overhead.
\Cref{fig:effect_of_k_topk_dynamics} shows the training dynamics and reveals where the differences arise. Top-$1$ exhibits unstable overlap growth, accompanied by sharp spikes in entropy and gradient norm. Top-$4$ is substantially more stable but still shows a late-stage dip. Top-$16$ and Top-$64$ remain smooth throughout.

Overall, these results suggest that the support size may not be a critical design choice for OPD, as long as the degenerate Top-$1$ setting is avoided. The reason sampled-token OPD works well despite using only one token per position is that it draws a different token at each step proportionally to the student's own distribution, providing unbiased coverage of the high-probability region across training. Top-$1$, by contrast, always selects the argmax token, thereby concentrating the reward on a single mode. Small policy changes can flip which token occupies rank 1, creating an unstable reward signal that does not average out over training. The failure of Top-$1$ is therefore not about using too few tokens, but about using a biased, mode-concentrated selection rule.

%% file: Sections/7_Related_Works.tex
\section{Related Work}

\paragraph{Knowledge Distillation.}
Knowledge distillation (KD)~\citep{hinton2015distilling} transfers knowledge from a large model to a smaller one by training a student network on the soft output distributions of a teacher.
For autoregressive sequence models,~\citet{kim2016sequence} extended this to sequence-level distillation by training students on teacher-generated outputs, establishing the dominant off-policy distillation baseline~\citep{sanh2019distilbert,jiao2020tinybert,wang2020minilm}.
In parallel, supervised fine-tuning (SFT) has been directly applied to improve performance on a variety of downstream tasks~\citep{JMLR:v25:23-0870,sanh2021multitask,wei2021finetuned}.
A fundamental limitation shared by all off-policy approaches is the train-inference distribution mismatch. The student is optimized on teacher-generated or reference sequences, but must generate from its own distribution at inference, which is an instance of the exposure bias~\citep{bengio2015scheduled} that accumulates errors over long generations.
This mismatch motivates shifting distillation to the student's own on-policy distribution, which is the central idea behind on-policy distillation.

\paragraph{On-Policy Distillation.}
MiniLLM~\citep{gu2023minillm} first formalized on-policy distillation (OPD) for LLMs under a reverse KL objective optimized via policy gradient, arguing that reverse KL's mode-seeking behavior prevents the student from spreading probability mass over regions the teacher considers unlikely.
GKD~\citep{agarwal2024policy} introduced a unified framework interpolating between on-policy and off-policy data across multiple divergences, demonstrating consistent gains over other KD baselines.
\citet{yang2026learning} later formalized OPD theoretically as a special case of dense KL-constrained RL, showing that the teacher's per-token log-ratio constitutes an implicit reward and that scaling this reward beyond its standard weight can push the student past the teacher's performance boundary.
OPD has since been adopted in industry post-training pipelines~\citep{yang2025qwen3,lu2025onpolicydistillation,zeng2026glm,xiao2026mimo,ko2026scaling,jin2026entropy,jang2026stable, fu2026revisiting,yang2026learning}, and extended to scalable self-distillation~\citep{hubotter2026reinforcement,zhao2026self,he2026far,shenfeld2026self,ye2026policy,sang2026crispcompressedreasoningiterative,kim2026does,ye2026online,yang2026self,li2026unifying,zhao2026selfdistillationmultitokenprediction,ding2026hdpo}, where a single model acts as its own teacher by conditioning on privileged information such as ground-truth solutions or execution feedback.
Despite this growing body of work, existing studies focus on demonstrating OPD's promise, such as dense rewards and mitigated exposure bias, across varied objectives, tasks, and teacher-student pairs, without systematically analyzing when or why OPD fails.

\paragraph{Capacity Gap and Distillability.}
A recurring observation in knowledge distillation is that large teacher-student capacity gaps can degrade or even reverse the benefit of distillation.
\citet{cho2019efficacy} demonstrate that distillation can hurt student performance when the teacher is substantially more capable, and~\citet{mirzadeh2020improved} propose an intermediate-sized teacher assistant to bridge the gap.
\citet{busbridge2025distillation} provide a quantitative treatment via distillation scaling laws, showing that student loss follows a power law as a function of teacher quality, student size, and data volume, identifying a U-shaped capacity regime where teacher over-capability degrades distillation efficiency.
For LLM reasoning,~\citet{li2025small} document a ``learnability gap'' showing that training small models on long chain-of-thought traces from strong reasoning teachers consistently underperforms simpler approaches, suggesting that the reasoning complexity of teacher outputs must be matched to student capacity.
These findings call for caution regarding the universality of distillation. However, the existing analyses have largely centered on off-policy knowledge distillation. In particular, the issues of capacity gap and distillability in OPD remain underexplored.

%% file: Sections/8_Conclusion.tex
\section{Conclusion and Future Work}

This work provides a systematic analysis of OPD, decomposing its success into two governing conditions: thinking-pattern consistency and the presence of genuinely new knowledge beyond what the student has seen during training.
When these conditions are unmet, off-policy cold start and teacher-aligned prompt selection provide effective remedies.
We also reveal a practical ceiling imposed by reward degradation over long trajectories.

\textbf{Future Work} \quad Building on our findings, we identify several directions for future research:
\begin{itemize}[leftmargin=12pt]
    \item \textbf{Beyond Mathematical Reasoning:} All experiments in this work are conducted on mathematical benchmarks. Whether the same conditions and token-level mechanisms govern OPD in other domains such as code and open-ended settings remains an important open question.
    \item \textbf{Impact of Pre-Training:} The “new knowledge” condition implicitly depends on differences in pre-training corpora, but isolating this factor is challenging. Current studies mainly rely on cross-family distillation (e.g., Qwen → LLaMA), which confounds data divergence with tokenizer mismatch and architectural differences, while controlled pre-training ablations remain prohibitively expensive. As a result, measuring the effect of pre-training data on OPD remains an open problem.
    \item \textbf{Self-Distillation Dynamics:} Recent work increasingly adopts self-distillation, where a single model serves as its own teacher given privileged information.
    Extending these insights to the self-distillation regime, where thinking-pattern consistency is guaranteed but knowledge novelty arises from privileged access rather than a separate teacher, is a natural next step.
    \item \textbf{Long-Horizon and Agentic Settings:} The trajectory-length ceiling revealed in \Cref{sec:discussion} motivates hybrid approaches that combine dense token-level supervision on short segments with sparse outcome-level rewards for longer horizons, as well as curriculum strategies that progressively extend the supervised horizon during training.
\end{itemize}

%% file: Sections/Appendix.tex
\appendix

\section{Details for \Cref{sec:phenomenology}}

\subsection{GRPO Training Details}
\label{app:grpo_details}

\paragraph{Base Model.}
We initialize GRPO training from Qwen3-4B-Base.

\paragraph{Training Dataset.}
We use the processed DAPO-Math-17K dataset for GRPO training. Specifically, each question is augmented with the following instruction:
\begin{tcolorbox}[title=\textbf{GRPO dataset template}, colback=CornflowerBlue!20, colframe=CornflowerBlue!90!Black]
\small
\{Question\} Please reason step by step, and put your final answer within \textbackslash boxed\{\}.
\end{tcolorbox}

\paragraph{Training and Evaluation Settings.}
We train the teacher model using GRPO. During training, we sample $n=8$ responses for each prompt. The maximum prompt length and maximum response length are set to 1,024 and 7,168 tokens, respectively. Training is conducted for one epoch on 8 A800 80G GPUs with a learning rate of $1\times10^{-6}$. We set both the student sampling temperature and the teacher temperature to 1.0, use a repetition penalty of 1.0, disable KL regularization, and adopt \texttt{token-mean} loss aggregation. The main hyperparameters are summarized in \Cref{tab:grpo_hparams}.

\begin{table}[!h]
\centering
\caption{Training hyperparameters of GRPO for Qwen3-4B-Base-GRPO.}
\label{tab:grpo_hparams}
\begin{tabular}{ll}
\toprule
\textbf{Hyper-parameter} & \textbf{Value} \\
\midrule
Base model & Qwen3-4B-Base \\
RL algorithm & GRPO \\
Training epochs & 1 \\
Train batch size & 64 \\
Micro batch size & 64 \\
Rollout $n$ & 8 \\
Maximum prompt length & 1,024 \\
Maximum response length & 7,168 \\
Validation max response length & 31,744 \\
Learning rate & $1 \times 10^{-6}$ \\
Temperature & 1.0 \\
Top-$p$ & 1.0 \\
KL regularization & 0.0 \\
Loss aggregation & \texttt{token-mean} \\
KL Coefficient & 0.0\\
\bottomrule
\end{tabular}
\end{table}

\subsection{Experimental Setup}
\label{app:experimental_setup}

Unless otherwise noted, all experiments use the default OPD hyperparameters listed in Table~\ref{tab:train_hparams_script}.

\begin{table}[t]
\centering
\caption{\textbf{Default hyperparameters for OPD.}}
\label{tab:train_hparams_script}

\begin{tabular}{lc}
\toprule
\textbf{Item} & \textbf{Value} \\
\midrule
Training temperature & 1.0 \\
Global batch size & 64 \\
Mini batch size & 64 \\
Rollout number & 4 \\
LogProb top-$K$ & 16 \\
Top-$K$ strategy & Student Top-$K$ \\
Top-$p$ & 1.0 \\
Max prompt length & 1024 \\
Max response length & 7168 \\
Learning rate & 1e-6 \\
Epoch & 1 \\
KL Coefficient & 0.0\\
\bottomrule
\end{tabular}
\end{table}

\subsection{Benchmark-wise breakdown of thinking-pattern compatibility}
\label{app:thinking_pattern_breakdown}

\begin{figure*}[t]
    \centering
    \includegraphics[width=\textwidth]{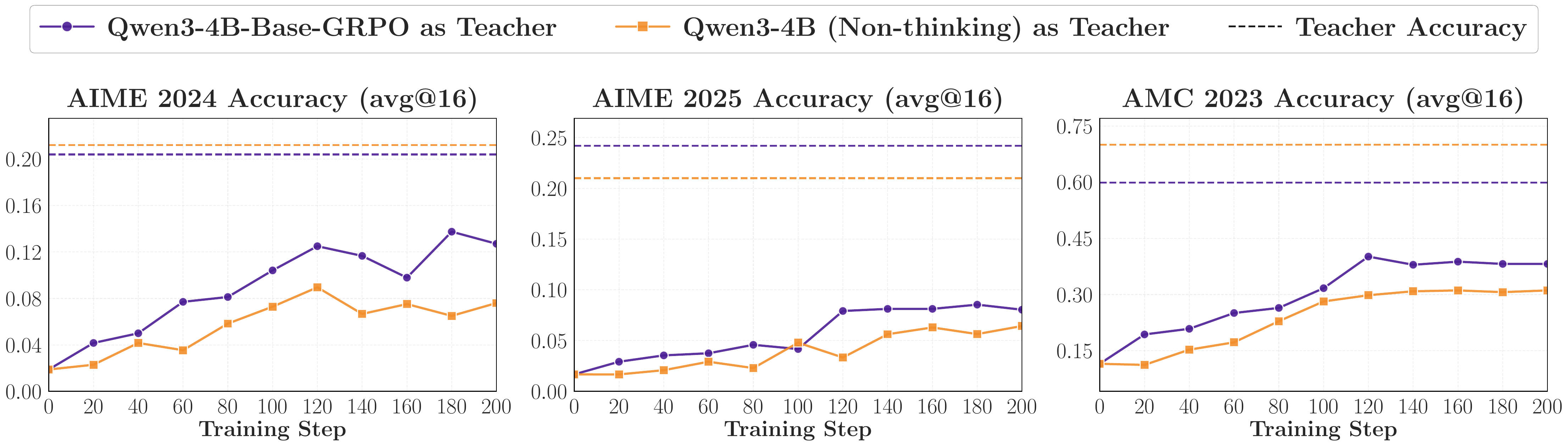}
    \caption{Benchmark-wise breakdown of the average validation accuracy shown in \Cref{fig:thinking_pattern_compatibility}. We report results on AIME 2024, AIME 2025, and AMC 2023 separately. Distillation from Qwen3-4B-Base-GRPO consistently matches or outperforms distillation from Qwen3-4B (Non-thinking) across the three benchmarks.}
    \label{fig:thinking_pattern_compatibility_breakdown}
\end{figure*}

To further unpack the averaged result in \Cref{fig:thinking_pattern_compatibility}, \Cref{fig:thinking_pattern_compatibility_breakdown} presents a benchmark-wise breakdown.
The advantage of distillation from Qwen3-4B-Base-GRPO is broadly consistent across datasets rather than being driven by a single benchmark.
The gap is more pronounced on AMC 2023 and AIME 2024, and smaller but still generally present on AIME 2025.
This per-benchmark view supports the interpretation that better early-stage thinking-pattern compatibility leads to better downstream distillation performance, and the loss from an early mismatch is not fully recovered later in training.

\section{Details for \Cref{sec:mechanism}}

\subsection{Additional Analysis of Token Overlap Mass}
\label{app:overlap_mass}

\begin{figure}[t]
    \centering
    \includegraphics[width=\linewidth]{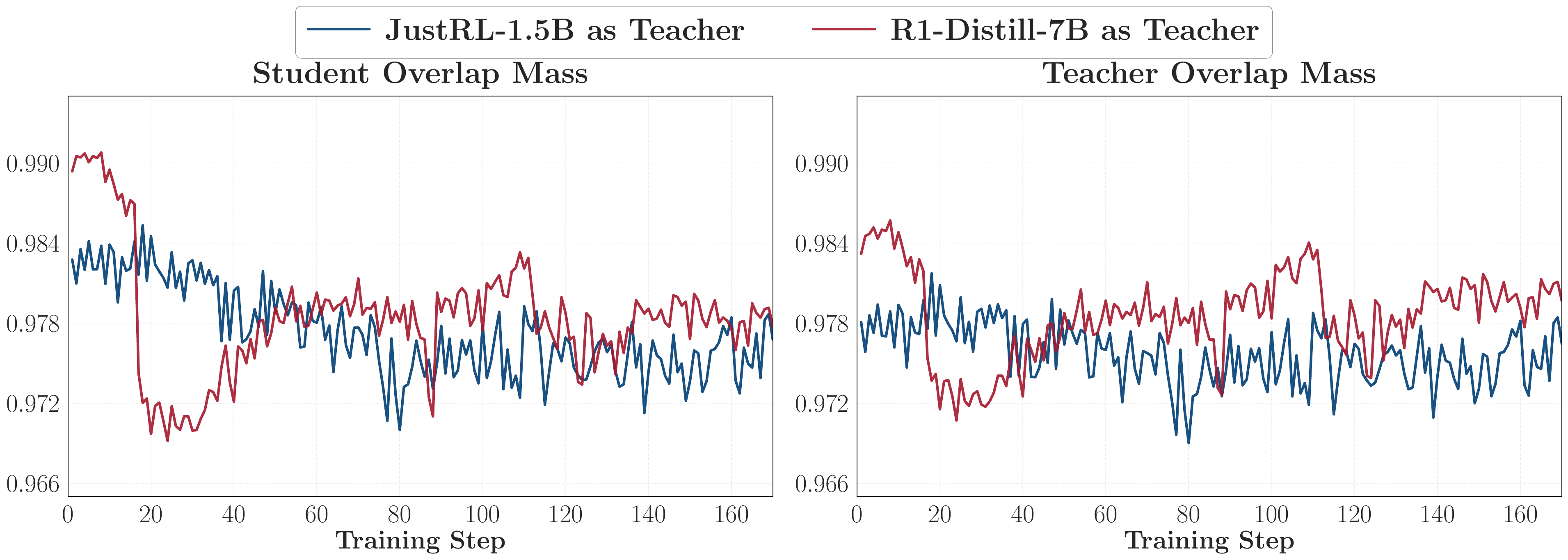}
    \caption{
    Probability mass assigned to overlap tokens during training.
    For both the student and teacher distributions, the overlap tokens consistently account for roughly 97\%--99\% of the total probability mass, indicating that the overlap is not only increasing at the set level but also dominates the probability distribution.
    }
    \label{fig:overlap-mass}
\end{figure}

To quantify how much probability mass each model assigns to the overlap top-$k$ region, we define:
\begin{equation}
\mathcal{M}_{\text{overlap-mass}}^{(p)}
=
\mathbb{E}_{t}\left[
  \sum_{v \in S_t^{(p)} \cap S_t^{(q)}} p_t(v)
\right],
\label{eq:metric_student_overlap_mass}
\end{equation}
and
\begin{equation}
\mathcal{M}_{\text{overlap-mass}}^{(q)}
=
\mathbb{E}_{t}\left[
  \sum_{v \in S_t^{(p)} \cap S_t^{(q)}} q_t(v)
\right],
\label{eq:metric_teacher_overlap_mass}
\end{equation}
which measure the fraction of total probability mass that the student and teacher, respectively, assign to the shared tokens in their top-$k$ sets. In our experiments, the overlap tokens carry $97\%$--$99\%$ of the total probability mass for both models throughout training, as shown in \Cref{fig:overlap-mass}.

\subsection{Auxiliary Optimization Dynamics}
\label{app:auxiliary_dynamics}

To complement the analysis in \Cref{sec:alignment_high_prob}, we report several additional optimization diagnostics for the same contrastive setting. Throughout this appendix, we fix the student to R1-Distill-1.5B and compare two teachers under the same Student Top-$k$ OPD training recipe: JustRL-1.5B, which yields a successful run, and R1-Distill-7B, which yields a failing run under otherwise matched conditions. These diagnostics are not intended as primary evidence; rather, they provide a complementary view of how the optimization signal differs between successful and failing OPD.

\begin{figure*}[t]
    \centering
    \includegraphics[width=\textwidth]{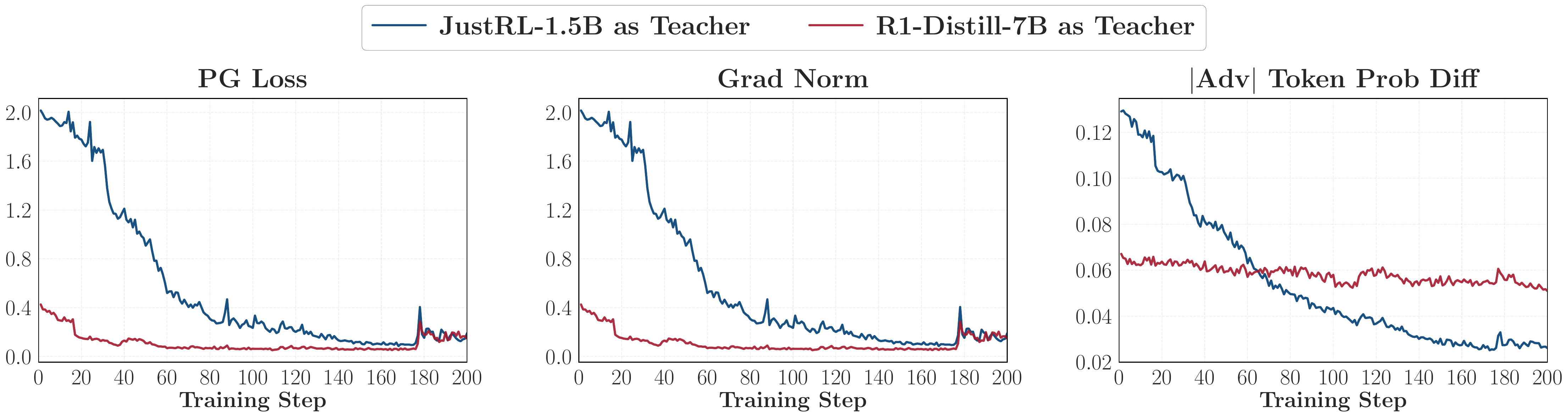}
    \caption{
    Auxiliary optimization diagnostics for the contrastive OPD setting in \Cref{sec:alignment_high_prob}, using R1-Distill-1.5B as the student and comparing JustRL-1.5B against R1-Distill-7B as the teacher. 
    \textbf{Left:} batch-averaged OPD training loss (\textit{PG Loss}) over training. 
    \textbf{Middle:} gradient norm over training. 
    \textbf{Right:} probability difference $p_t(v)-q_t(v)$ measured on the token with the largest absolute advantage. 
    The successful run exhibits a large reduction in optimization loss, sustained gradient magnitude, and a steady decrease in extreme-token probability mismatch. By contrast, the failing run starts with and maintains much weaker gradients, and its extreme-token probability discrepancy remains noticeably larger throughout training.
    }
    \label{fig:auxiliary_dynamics}
\end{figure*}

\paragraph{Diagnostics.}
We monitor three additional quantities. The first is the batch-averaged OPD training loss, denoted as \emph{PG Loss} in \Cref{fig:auxiliary_dynamics}. The second is the gradient norm, which measures the overall magnitude of the update signal reaching the student. The third is the probability difference $p_t(v)-q_t(v)$ on the token with the largest absolute advantage, which tracks whether the student can reduce the most pronounced local disagreement with the teacher on the tokens that carry the strongest optimization signal. Together, these metrics help distinguish between successful and failing OPD: in the former, the student receives a usable signal and progressively reduces mismatch, whereas in the latter, the signal is too weak or too poorly aligned to drive substantial improvement.

\paragraph{Results.}
The trends in \Cref{fig:auxiliary_dynamics} are consistent with the main conclusion of \Cref{sec:alignment_high_prob}. First, the successful run with JustRL-1.5B shows a pronounced reduction in training loss over the course of optimization. Starting from a much larger initial mismatch, the loss decreases steadily for most of training before flattening at a low value. By contrast, the failing run with R1-Distill-7B begins with a much smaller loss and changes only modestly thereafter. This pattern suggests that the smaller loss in the failing run does not indicate better optimization. Rather, it reflects a weak teacher-induced training signal from the outset, which remains too small to drive substantial policy improvement.

Second, the gradient norm shows an even clearer separation between the two runs. In the successful run, the gradient norm is initially large and remains substantial through a long portion of training, indicating that the student continues to receive a meaningful corrective signal. In the failing run, the gradient norm is consistently much smaller, with only limited variation over time. Thus, even though optimization proceeds under the same algorithm and training budget, the student trained against R1-Distill-7B experiences a much weaker update signal. This observation is consistent with the finding that failure is associated with poor alignment on high-probability tokens: when the student does not meaningfully enter the teacher-supported region, the resulting gradients remain weak.

Third, the right panel shows that the successful run steadily reduces the probability discrepancy on the token with the largest absolute advantage, whereas the failing run maintains a noticeably larger gap throughout training. In other words, when OPD succeeds, the student progressively corrects the local mistakes that matter most under the teacher-induced advantage signal. When OPD fails, these high-advantage discrepancies persist rather than being resolved. This is again consistent with the interpretation that the decisive signal in OPD lies on a small set of high-probability, high-advantage tokens, and failure occurs when the student cannot effectively exploit that signal.

Taken together, these auxiliary dynamics reinforce the interpretation developed in \Cref{sec:alignment_high_prob}. Successful OPD is characterized not only by increasing overlap on high-probability tokens, but also by a training regime in which the student receives gradients of sufficient magnitude to reduce the most important local distributional mismatches. In contrast, failing OPD is accompanied by weak gradients, limited loss reduction, and persistent disagreement on the tokens with the strongest advantage signal. While these diagnostics are supportive rather than central, they provide an optimization-level view that is fully consistent with that the useful learning signal of OPD is concentrated on high-probability tokens at student-visited states, and training degrades when that signal is too weak or too misaligned to drive effective updates.

\subsection{Cross-Model Validation of High-Probability-Token Alignment}
\label{app:cross_model_validation}

We further test whether the phenomenon in \Cref{sec:alignment_high_prob} generalizes to another model pair.
Here we fix the student model to R1-Distill-7B and choose Skywork-OR1-Math-7B and DeepSeek-R1-Distill-Qwen-14B (R1-Distill-14B) as teachers, using the same training and evaluation setup as in \Cref{sec:alignment_high_prob}.

\begin{figure*}[t]
    \centering
    \includegraphics[width=\textwidth]{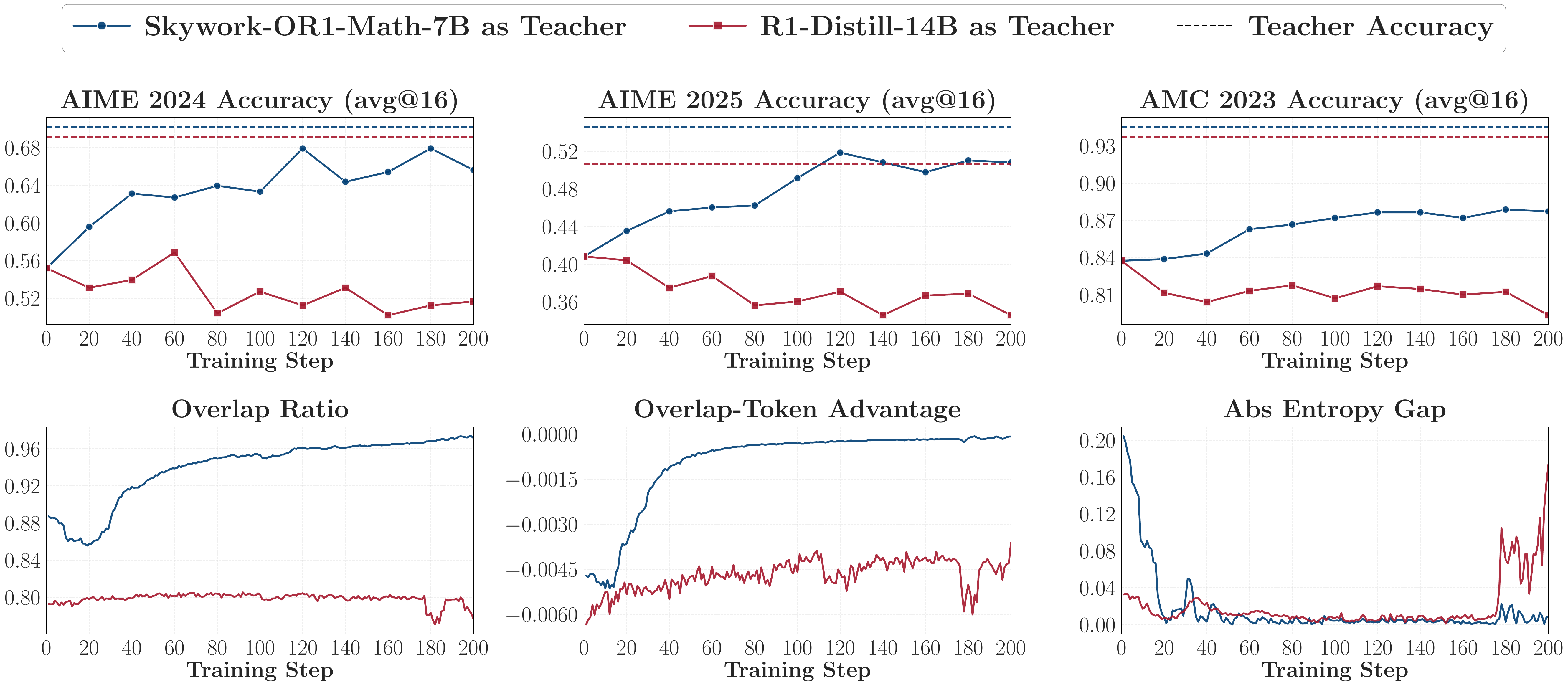}
    \caption{
    Cross-model validation with a fixed student (R1-Distill-7B) and two teachers.
    \textbf{Top:} avg@16 accuracy on AIME 2024, AIME 2025, and AMC 2023.
    \textbf{Bottom:} overlap ratio, overlap-token advantage, and absolute entropy gap over training.
    The successful run is again accompanied by increasing high-probability-token alignment, while the stagnating run is not.
    }
    \label{fig:alignment_cross_model}
\end{figure*}

\paragraph{Results.}
\Cref{fig:alignment_cross_model} shows the same pattern as \Cref{fig:alignment_main}.
With Skywork-OR1-Math-7B as the teacher, distillation improves student performance and is accompanied by steadily increasing overlap ratio, overlap-token advantage approaching zero, and a small entropy gap.
In contrast, with R1-Distill-14B as the teacher, training shows little improvement and the alignment metrics remain poor or unstable.
This provides additional evidence that successful OPD consistently coincides with the emergence of high-probability-token alignment at student-visited states.

\section{Details for \Cref{sec:recipe}}

\subsection{Cold-Start Distillation Details}
\label{app:coldstart_details}

\paragraph{Offline teacher rollout.}
To construct the cold-start SFT data, we sample 200K math prompts from the math subset of OpenThoughts3-1.2M~\citep{guha2025openthoughts} and use Qwen3-4B (Non-thinking) to generate one offline response for each prompt. For each prompt, we use the following template:
\begin{tcolorbox}[title=\textbf{Teacher rollout template}, colback=CornflowerBlue!20, colframe=CornflowerBlue!90!Black]
\small
\{Question\} Please reason step by step, and put your final answer within \textbackslash boxed\{\}.
\end{tcolorbox}
We decode with temperature $0.7$, top-$p=0.95$, top-$k=-1$, and a maximum generation length of 12{,}288 tokens. After generation, we filter out incomplete responses (e.g., truncated outputs that do not finish properly) and degenerate repetitive responses. The remaining prompt-response pairs are used as the supervised distillation corpus for training the student.

\paragraph{Student SFT.}
Starting from Qwen3-1.7B-Base, we perform full-parameter SFT on the filtered 200K teacher-generated samples using the LLaMA-Factory framework~\citep{zheng2024llamafactory}, yielding Qwen3-1.7B-SFT. We summarize the detailed hyperparameters in Table~\ref{tab:coldstart_sft_hparams}.

\begin{table}[t]
\centering
\caption{SFT hyperparameters for cold-start distillation from Qwen3-4B (Non-thinking) to Qwen3-1.7B-Base.}
\label{tab:coldstart_sft_hparams}
\begin{tabular}{ll}
\toprule
\textbf{Hyper-parameter} & \textbf{Value} \\
\midrule
Student model & Qwen3-1.7B-Base \\
Training objective & Full-parameter SFT \\
Template & \texttt{qwen3} \\
Training epochs & 1 \\
Sequence length & 14{,}336 \\
Per-device batch size & 8 \\
Gradient accumulation steps & 1 \\
Learning rate & $1 \times 10^{-5}$ \\
LR scheduler & Cosine \\
Warmup ratio & 0.05 \\
Precision & BF16 \\
\bottomrule
\end{tabular}
\end{table}

\subsection{Additional Analysis of Overlap Mass}
\label{app:add_analysis_of_overlap_mass}

\begin{figure}[t]
    \centering
    \includegraphics[width=\linewidth]{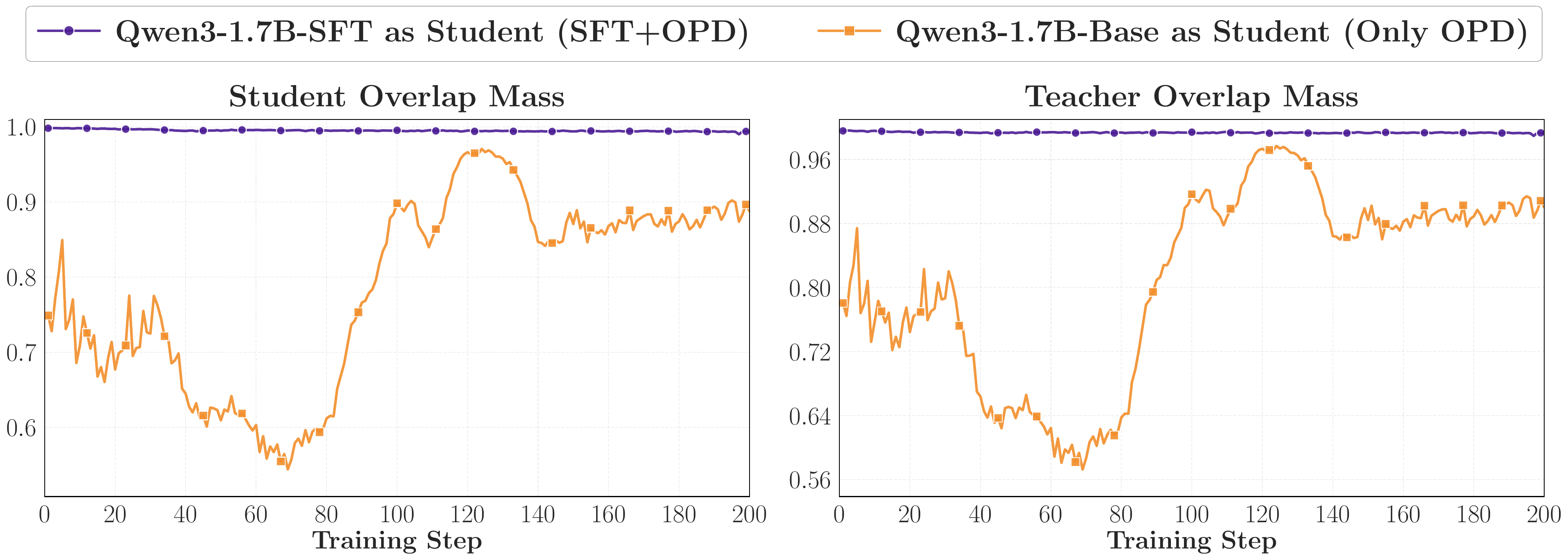}
    \caption{Student overlap mass and teacher overlap mass during training for SFT-initialized and base-initialized students.}
    \label{fig:overlap_mass}
\end{figure}

To better understand why the base-initialized student can occasionally exhibit a comparable or even slightly better Overlap-Token Advantage while still underperforming overall, we further examine the probability mass covered by the overlap set from both the student and teacher sides. As shown in Figure~\ref{fig:overlap_mass}, the SFT-initialized student maintains both student overlap mass and teacher overlap mass at consistently high levels throughout training. This indicates that the overlap tokens cover most of the high-probability regions of both the student and teacher distributions, suggesting a strong and stable alignment from the beginning of OPD. In contrast, the base-initialized student exhibits substantially lower and more unstable overlap mass, especially in the early stage of training. 

This analysis helps explain why Overlap-Token Advantage alone can sometimes be misleading. Since it is averaged only over overlap tokens, it can appear relatively favorable even when the overlap set itself misses substantial high-probability teacher tokens. Overlap mass complements this view by revealing whether the shared support actually covers the most important parts of the two distributions. From this perspective, the SFT cold start leads to a substantially better and more stable match between student and teacher.

\subsection{Deduplication Details for the DeepMath Subset}
\label{app:deepmath_dedup}

For the cross-size setting, we construct a DeepMath subset deduplicated against DAPO-Math-17K in order to compare prompts aligned with the teacher's RL post-training data against prompts that are only in-domain.

Our deduplication is performed in two stages: exact-match deduplication and semantic deduplication.

\paragraph{Question extraction.}
For both DAPO-Math-17K and DeepMath, we extract the question content and remove the instruction suffix in the prompt, so that deduplication is performed based on the question text alone.

\paragraph{Stage 1: Exact-match deduplication.}
We collect all extracted DAPO-Math-17K questions into a set and remove any DeepMath example whose extracted question exactly matches one of the DAPO questions.

\paragraph{Stage 2: Semantic deduplication.}
To further remove near-duplicate prompts, we encode both DAPO-Math-17K and DeepMath questions using the sentence embedding model all-mpnet-base-v2~\citep{reimers2019sentencebert}.
We L2-normalize the embeddings and build a FAISS inner-product index over the DAPO embeddings, so that the inner product corresponds to cosine similarity.
For each DeepMath question, we retrieve its top-1 nearest neighbor in DAPO-Math-17K.
If the cosine similarity to the nearest DAPO question is at least $0.6$, we mark the DeepMath example as a semantic duplicate and remove it.

\paragraph{Final retained subset.}
We remove any DeepMath example flagged by either exact-match or semantic deduplication.
The resulting subset is in-domain but deduplicated against DAPO-Math-17K, enabling a controlled comparison between prompts that overlap with the teacher's post-training data and prompts that are only in-domain.

\begin{figure*}[t]
    \centering
    \includegraphics[width=\textwidth]{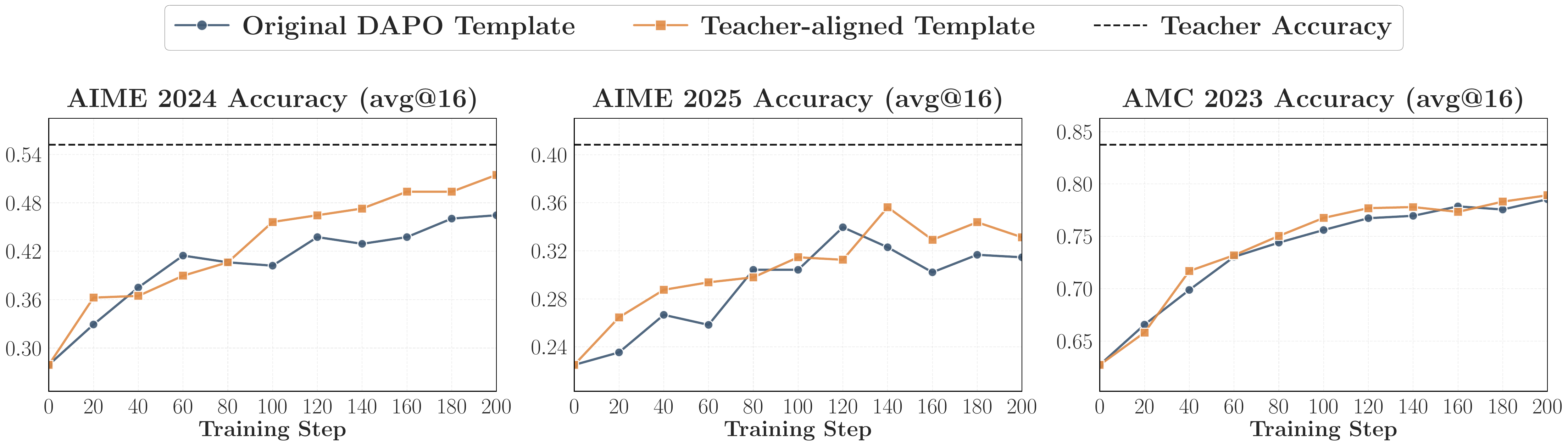}
    \caption{Benchmark-wise breakdown of the average validation accuracy shown in \Cref{fig:teacher_posttrain_data_same}. Using the teacher-aligned template consistently matches or outperforms the original DAPO template across the three benchmarks.}
    \label{fig:teacher_posttrain_data_same_breakdown}
\end{figure*}

\subsection{Benchmark-wise breakdown of prompt-template alignment}
\label{app:teacher_template_breakdown}

To further unpack the averaged result in \Cref{fig:teacher_posttrain_data_same}, \Cref{fig:teacher_posttrain_data_same_breakdown} presents a benchmark-wise breakdown.
The teacher-aligned template yields broadly consistent improvements across datasets, with larger gains on the two AIME sets and a smaller but still positive effect on AMC 2023.
It also allows the student to recover a larger fraction of the teacher's performance, increasing from roughly 80\% to roughly 85\%.
Together with the overlap-ratio result in \Cref{sec:teacher_posttrain_data}, this suggests that prompt-template alignment improves OPD by making the student's generated states more compatible with the teacher.

\section{Details for \Cref{sec:discussion}}

\subsection{Teacher entropy by output position}
\label{appendix:teacher_entropy_by_position}

\begin{figure*}[t]
    \centering
    \includegraphics[width=\textwidth]{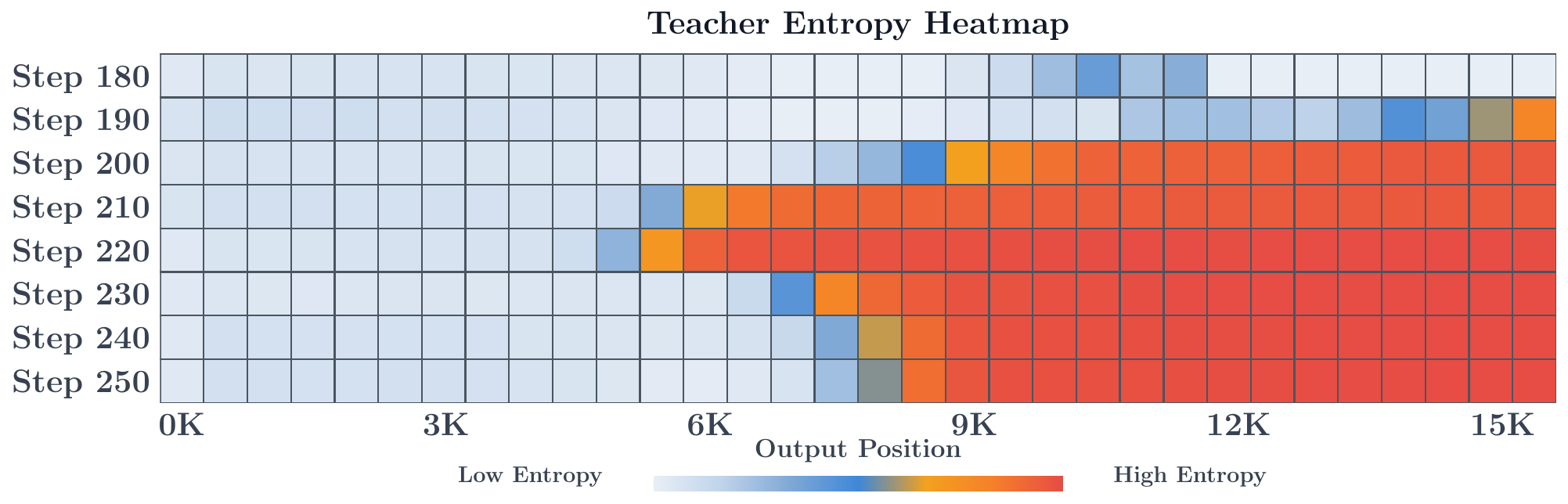}
    \caption{
    Average teacher entropy across decoding positions during OPD training with 15K max response length, measured on student-generated trajectories from Step 180 to Step 250. Elevated entropy first emerges in the suffix and gradually propagates toward earlier output positions over training.
    }
    \label{fig:teacher_entropy_by_position_appendix}
\end{figure*}

To complement the student entropy analysis in \Cref{sec:reward_noise}, we also visualize teacher entropy as a function of output position across training steps under the 15$K$ max response length setting (see \Cref{fig:teacher_entropy_by_position_appendix}). Similar to the student, teacher entropy first increases at later decoding positions and then progressively propagates toward earlier tokens over training.

%% file: opd.bbl
\begin{thebibliography}{47}
\providecommand{\natexlab}[1]{#1}
\providecommand{\url}[1]{\texttt{#1}}
\expandafter\ifx\csname urlstyle\endcsname\relax
  \providecommand{\doi}[1]{doi: #1}\else
  \providecommand{\doi}{doi: \begingroup \urlstyle{rm}\Url}\fi

\bibitem[Agarwal et~al.(2024)Agarwal, Vieillard, Zhou, Stanczyk, Garea, Geist, and Bachem]{agarwal2024policy}
Rishabh Agarwal, Nino Vieillard, Yongchao Zhou, Piotr Stanczyk, Sabela~Ramos Garea, Matthieu Geist, and Olivier Bachem.
\newblock On-policy distillation of language models: Learning from self-generated mistakes.
\newblock In \emph{The twelfth international conference on learning representations}, 2024.

\bibitem[Balunovi{\'c} et~al.(2025)Balunovi{\'c}, Dekoninck, Petrov, Jovanovi{\'c}, and Vechev]{balunovic2025matharena}
Mislav Balunovi{\'c}, Jasper Dekoninck, Ivo Petrov, Nikola Jovanovi{\'c}, and Martin Vechev.
\newblock Matharena: Evaluating llms on uncontaminated math competitions.
\newblock \emph{arXiv preprint arXiv:2505.23281}, 2025.

\bibitem[Bengio et~al.(2015)Bengio, Vinyals, Jaitly, and Shazeer]{bengio2015scheduled}
Samy Bengio, Oriol Vinyals, Navdeep Jaitly, and Noam Shazeer.
\newblock Scheduled sampling for sequence prediction with recurrent neural networks.
\newblock \emph{Advances in neural information processing systems}, 28, 2015.

\bibitem[Busbridge et~al.(2025)Busbridge, Shidani, Weers, Ramapuram, Littwin, and Webb]{busbridge2025distillation}
Dan Busbridge, Amitis Shidani, Floris Weers, Jason Ramapuram, Etai Littwin, and Russ Webb.
\newblock Distillation scaling laws.
\newblock \emph{arXiv preprint arXiv:2502.08606}, 2025.

\bibitem[Cho and Hariharan(2019)]{cho2019efficacy}
Jang~Hyun Cho and Bharath Hariharan.
\newblock On the efficacy of knowledge distillation.
\newblock In \emph{Proceedings of the IEEE/CVF international conference on computer vision}, pages 4794--4802, 2019.

\bibitem[Chung et~al.(2024)Chung, Hou, Longpre, Zoph, Tay, Fedus, Li, Wang, Dehghani, Brahma, Webson, Gu, Dai, Suzgun, Chen, Chowdhery, Castro-Ros, Pellat, Robinson, Valter, Narang, Mishra, Yu, Zhao, Huang, Dai, Yu, Petrov, Chi, Dean, Devlin, Roberts, Zhou, Le, and Wei]{JMLR:v25:23-0870}
Hyung~Won Chung, Le~Hou, Shayne Longpre, Barret Zoph, Yi~Tay, William Fedus, Yunxuan Li, Xuezhi Wang, Mostafa Dehghani, Siddhartha Brahma, Albert Webson, Shixiang~Shane Gu, Zhuyun Dai, Mirac Suzgun, Xinyun Chen, Aakanksha Chowdhery, Alex Castro-Ros, Marie Pellat, Kevin Robinson, Dasha Valter, Sharan Narang, Gaurav Mishra, Adams Yu, Vincent Zhao, Yanping Huang, Andrew Dai, Hongkun Yu, Slav Petrov, Ed~H. Chi, Jeff Dean, Jacob Devlin, Adam Roberts, Denny Zhou, Quoc~V. Le, and Jason Wei.
\newblock Scaling instruction-finetuned language models.
\newblock \emph{Journal of Machine Learning Research}, 25\penalty0 (70):\penalty0 1--53, 2024.
\newblock URL \url{http://jmlr.org/papers/v25/23-0870.html}.

\bibitem[Ding(2026)]{ding2026hdpo}
Ken Ding.
\newblock Hdpo: Hybrid distillation policy optimization via privileged self-distillation.
\newblock \emph{arXiv preprint arXiv:2603.23871}, 2026.

\bibitem[Fu et~al.(2026)Fu, Huang, Jiang, Zhu, and Zhao]{fu2026revisiting}
Yuqian Fu, Haohuan Huang, Kaiwen Jiang, Yuanheng Zhu, and Dongbin Zhao.
\newblock Revisiting on-policy distillation: Empirical failure modes and simple fixes.
\newblock \emph{arXiv preprint arXiv:2603.25562}, 2026.

\bibitem[Gu et~al.(2023)Gu, Dong, Wei, and Huang]{gu2023minillm}
Yuxian Gu, Li~Dong, Furu Wei, and Minlie Huang.
\newblock Minillm: Knowledge distillation of large language models.
\newblock \emph{arXiv preprint arXiv:2306.08543}, 2023.

\bibitem[Guha et~al.(2025)Guha, Marten, Keh, Raoof, Smyrnis, Bansal, Nezhurina, Mercat, Vu, Sprague, et~al.]{guha2025openthoughts}
Etash Guha, Ryan Marten, Sedrick Keh, Negin Raoof, Georgios Smyrnis, Hritik Bansal, Marianna Nezhurina, Jean Mercat, Trung Vu, Zayne Sprague, et~al.
\newblock Openthoughts: Data recipes for reasoning models.
\newblock \emph{arXiv preprint arXiv:2506.04178}, 2025.

\bibitem[Guo et~al.(2025)Guo, Yang, Zhang, Song, Wang, Zhu, Xu, Zhang, Ma, Bi, et~al.]{guo2025deepseek}
Daya Guo, Dejian Yang, Haowei Zhang, Junxiao Song, Peiyi Wang, Qihao Zhu, Runxin Xu, Ruoyu Zhang, Shirong Ma, Xiao Bi, et~al.
\newblock Deepseek-r1 incentivizes reasoning in llms through reinforcement learning.
\newblock \emph{Nature}, 645\penalty0 (8081):\penalty0 633--638, 2025.

\bibitem[He et~al.(2025{\natexlab{a}})He, Qu, Liu, Chen, Zuo, Qian, Zhang, Chen, Xiao, Cui, et~al.]{he2025justrl}
Bingxiang He, Zekai Qu, Zeyuan Liu, Yinghao Chen, Yuxin Zuo, Cheng Qian, Kaiyan Zhang, Weize Chen, Chaojun Xiao, Ganqu Cui, et~al.
\newblock Justrl: Scaling a 1.5 b llm with a simple rl recipe.
\newblock \emph{arXiv preprint arXiv:2512.16649}, 2025{\natexlab{a}}.

\bibitem[He et~al.(2026)He, Zuo, Liu, Zhao, Fu, Yang, Qian, Zhang, Fan, Cui, et~al.]{he2026far}
Bingxiang He, Yuxin Zuo, Zeyuan Liu, Shangziqi Zhao, Zixuan Fu, Junlin Yang, Cheng Qian, Kaiyan Zhang, Yuchen Fan, Ganqu Cui, et~al.
\newblock How far can unsupervised rlvr scale llm training?
\newblock \emph{arXiv preprint arXiv:2603.08660}, 2026.

\bibitem[He et~al.(2025{\natexlab{b}})He, Liu, Liu, Yan, Wang, Cheng, Zhang, Zhang, Xu, Shen, et~al.]{he2025skywork}
Jujie He, Jiacai Liu, Chris~Yuhao Liu, Rui Yan, Chaojie Wang, Peng Cheng, Xiaoyu Zhang, Fuxiang Zhang, Jiacheng Xu, Wei Shen, et~al.
\newblock Skywork open reasoner 1 technical report.
\newblock \emph{arXiv preprint arXiv:2505.22312}, 2025{\natexlab{b}}.

\bibitem[He et~al.(2025{\natexlab{c}})He, Liang, Xu, Liu, Chen, Wang, Song, Yu, Liang, Wang, et~al.]{he2025deepmath}
Zhiwei He, Tian Liang, Jiahao Xu, Qiuzhi Liu, Xingyu Chen, Yue Wang, Linfeng Song, Dian Yu, Zhenwen Liang, Wenxuan Wang, et~al.
\newblock Deepmath-103k: A large-scale, challenging, decontaminated, and verifiable mathematical dataset for advancing reasoning.
\newblock \emph{arXiv preprint arXiv:2504.11456}, 2025{\natexlab{c}}.

\bibitem[Hinton et~al.(2015)Hinton, Vinyals, and Dean]{hinton2015distilling}
Geoffrey Hinton, Oriol Vinyals, and Jeff Dean.
\newblock Distilling the knowledge in a neural network.
\newblock \emph{arXiv preprint arXiv:1503.02531}, 2015.

\bibitem[H{\"u}botter et~al.(2026)H{\"u}botter, L{\"u}beck, Behric, Baumann, Bagatella, Marta, Hakimi, Shenfeld, Buening, Guestrin, et~al.]{hubotter2026reinforcement}
Jonas H{\"u}botter, Frederike L{\"u}beck, Lejs Behric, Anton Baumann, Marco Bagatella, Daniel Marta, Ido Hakimi, Idan Shenfeld, Thomas~Kleine Buening, Carlos Guestrin, et~al.
\newblock Reinforcement learning via self-distillation.
\newblock \emph{arXiv preprint arXiv:2601.20802}, 2026.

\bibitem[Jang et~al.(2026)Jang, Yeom, Yeo, Lim, and Kim]{jang2026stable}
Ijun Jang, Jewon Yeom, Juan Yeo, Hyunggu Lim, and Taesup Kim.
\newblock Stable on-policy distillation through adaptive target reformulation.
\newblock \emph{arXiv preprint arXiv:2601.07155}, 2026.

\bibitem[Jiao et~al.(2020)Jiao, Yin, Shang, Jiang, Chen, Li, Wang, and Liu]{jiao2020tinybert}
Xiaoqi Jiao, Yichun Yin, Lifeng Shang, Xin Jiang, Xiao Chen, Linlin Li, Fang Wang, and Qun Liu.
\newblock Tinybert: Distilling bert for natural language understanding.
\newblock In \emph{Findings of the association for computational linguistics: EMNLP 2020}, pages 4163--4174, 2020.

\bibitem[Jin et~al.(2026)Jin, Min, Yang, Kadhe, Zhou, Wei, Baracaldo, and Lee]{jin2026entropy}
Woogyeol Jin, Taywon Min, Yongjin Yang, Swanand~Ravindra Kadhe, Yi~Zhou, Dennis Wei, Nathalie Baracaldo, and Kimin Lee.
\newblock Entropy-aware on-policy distillation of language models.
\newblock \emph{arXiv preprint arXiv:2603.07079}, 2026.

\bibitem[Kim et~al.(2026)Kim, Luo, Kim, Lee, Kim, Jeon, Li, and Yang]{kim2026does}
Jeonghye Kim, Xufang Luo, Minbeom Kim, Sangmook Lee, Dohyung Kim, Jiwon Jeon, Dongsheng Li, and Yuqing Yang.
\newblock Why does self-distillation (sometimes) degrade the reasoning capability of llms?
\newblock \emph{arXiv preprint arXiv:2603.24472}, 2026.

\bibitem[Kim and Rush(2016)]{kim2016sequence}
Yoon Kim and Alexander~M Rush.
\newblock Sequence-level knowledge distillation.
\newblock In \emph{Proceedings of the 2016 conference on empirical methods in natural language processing}, pages 1317--1327, 2016.

\bibitem[Ko et~al.(2026)Ko, Abdali, Kim, Chen, and Cameron]{ko2026scaling}
Jongwoo Ko, Sara Abdali, Young~Jin Kim, Tianyi Chen, and Pashmina Cameron.
\newblock Scaling reasoning efficiently via relaxed on-policy distillation.
\newblock \emph{arXiv preprint arXiv:2603.11137}, 2026.

\bibitem[Li et~al.(2026)Li, Yang, Fang, Song, Zheng, Guo, Zhang, Wang, and Chua]{li2026unifying}
Gengsheng Li, Tianyu Yang, Junfeng Fang, Mingyang Song, Mao Zheng, Haiyun Guo, Dan Zhang, Jinqiao Wang, and Tat-Seng Chua.
\newblock Unifying group-relative and self-distillation policy optimization via sample routing.
\newblock \emph{arXiv preprint arXiv:2604.02288}, 2026.

\bibitem[Li et~al.(2024)Li, Beeching, Tunstall, Lipkin, Soletskyi, Huang, Rasul, Yu, Jiang, Shen, et~al.]{li2024numinamath}
Jia Li, Edward Beeching, Lewis Tunstall, Ben Lipkin, Roman Soletskyi, Shengyi Huang, Kashif Rasul, Longhui Yu, Albert~Q Jiang, Ziju Shen, et~al.
\newblock Numinamath: The largest public dataset in ai4maths with 860k pairs of competition math problems and solutions.
\newblock \emph{Hugging Face repository}, 13:\penalty0 9, 2024.

\bibitem[Li et~al.(2025)Li, Yue, Xu, Jiang, Niu, Lin, Ramasubramanian, and Poovendran]{li2025small}
Yuetai Li, Xiang Yue, Zhangchen Xu, Fengqing Jiang, Luyao Niu, Bill~Yuchen Lin, Bhaskar Ramasubramanian, and Radha Poovendran.
\newblock Small models struggle to learn from strong reasoners.
\newblock In \emph{Findings of the Association for Computational Linguistics: ACL 2025}, pages 25366--25394, 2025.

\bibitem[Lu and Lab(2025)]{lu2025onpolicydistillation}
Kevin Lu and Thinking~Machines Lab.
\newblock On-policy distillation.
\newblock \emph{Thinking Machines Lab: Connectionism}, 2025.
\newblock \doi{10.64434/tml.20251026}.
\newblock https://thinkingmachines.ai/blog/on-policy-distillation.

\bibitem[Mirzadeh et~al.(2020)Mirzadeh, Farajtabar, Li, Levine, Matsukawa, and Ghasemzadeh]{mirzadeh2020improved}
Seyed~Iman Mirzadeh, Mehrdad Farajtabar, Ang Li, Nir Levine, Akihiro Matsukawa, and Hassan Ghasemzadeh.
\newblock Improved knowledge distillation via teacher assistant.
\newblock In \emph{Proceedings of the AAAI conference on artificial intelligence}, volume~34, pages 5191--5198, 2020.

\bibitem[Reimers and Gurevych(2019)]{reimers2019sentencebert}
Nils Reimers and Iryna Gurevych.
\newblock Sentence-bert: Sentence embeddings using siamese bert-networks.
\newblock In \emph{Proceedings of the 2019 Conference on Empirical Methods in Natural Language Processing}, 2019.
\newblock URL \url{http://arxiv.org/abs/1908.10084}.

\bibitem[Sang et~al.(2026)Sang, Xu, Zhou, He, Wang, and Sun]{sang2026crispcompressedreasoningiterative}
Hejian Sang, Yuanda Xu, Zhengze Zhou, Ran He, Zhipeng Wang, and Jiachen Sun.
\newblock Crisp: Compressed reasoning via iterative self-policy distillation, 2026.
\newblock URL \url{https://arxiv.org/abs/2603.05433}.

\bibitem[Sanh et~al.(2019)Sanh, Debut, Chaumond, and Wolf]{sanh2019distilbert}
Victor Sanh, Lysandre Debut, Julien Chaumond, and Thomas Wolf.
\newblock Distilbert, a distilled version of bert: smaller, faster, cheaper and lighter.
\newblock \emph{arXiv preprint arXiv:1910.01108}, 2019.

\bibitem[Sanh et~al.(2021)Sanh, Webson, Raffel, Bach, Sutawika, Alyafeai, Chaffin, Stiegler, Raja, Dey, et~al.]{sanh2021multitask}
Victor Sanh, Albert Webson, Colin Raffel, Stephen Bach, Lintang Sutawika, Zaid Alyafeai, Antoine Chaffin, Arnaud Stiegler, Arun Raja, Manan Dey, et~al.
\newblock Multitask prompted training enables zero-shot task generalization.
\newblock In \emph{International Conference on Learning Representations}, 2021.

\bibitem[Shao et~al.(2024)Shao, Wang, Zhu, Xu, Song, Bi, Zhang, Zhang, Li, Wu, et~al.]{shao2024deepseekmath}
Zhihong Shao, Peiyi Wang, Qihao Zhu, Runxin Xu, Junxiao Song, Xiao Bi, Haowei Zhang, Mingchuan Zhang, YK~Li, Yang Wu, et~al.
\newblock Deepseekmath: Pushing the limits of mathematical reasoning in open language models.
\newblock \emph{arXiv preprint arXiv:2402.03300}, 2024.

\bibitem[Shenfeld et~al.(2026)Shenfeld, Damani, H{\"u}botter, and Agrawal]{shenfeld2026self}
Idan Shenfeld, Mehul Damani, Jonas H{\"u}botter, and Pulkit Agrawal.
\newblock Self-distillation enables continual learning.
\newblock \emph{arXiv preprint arXiv:2601.19897}, 2026.

\bibitem[Wang et~al.(2020)Wang, Wei, Dong, Bao, Yang, and Zhou]{wang2020minilm}
Wenhui Wang, Furu Wei, Li~Dong, Hangbo Bao, Nan Yang, and Ming Zhou.
\newblock Minilm: Deep self-attention distillation for task-agnostic compression of pre-trained transformers.
\newblock \emph{Advances in neural information processing systems}, 33:\penalty0 5776--5788, 2020.

\bibitem[Wei et~al.(2021)Wei, Bosma, Zhao, Guu, Yu, Lester, Du, Dai, and Le]{wei2021finetuned}
Jason Wei, Maarten Bosma, Vincent Zhao, Kelvin Guu, Adams~Wei Yu, Brian Lester, Nan Du, Andrew~M Dai, and Quoc~V Le.
\newblock Finetuned language models are zero-shot learners.
\newblock In \emph{International Conference on Learning Representations}, 2021.

\bibitem[Xiao et~al.(2026)Xiao, Xia, Yang, Gao, Shen, Zhang, He, Lou, Luo, Wang, et~al.]{xiao2026mimo}
Bangjun Xiao, Bingquan Xia, Bo~Yang, Bofei Gao, Bowen Shen, Chen Zhang, Chenhong He, Chiheng Lou, Fuli Luo, Gang Wang, et~al.
\newblock Mimo-v2-flash technical report.
\newblock \emph{arXiv preprint arXiv:2601.02780}, 2026.

\bibitem[Yang et~al.(2025)Yang, Li, Yang, Zhang, Hui, Zheng, Yu, Gao, Huang, Lv, et~al.]{yang2025qwen3}
An~Yang, Anfeng Li, Baosong Yang, Beichen Zhang, Binyuan Hui, Bo~Zheng, Bowen Yu, Chang Gao, Chengen Huang, Chenxu Lv, et~al.
\newblock Qwen3 technical report.
\newblock \emph{arXiv preprint arXiv:2505.09388}, 2025.

\bibitem[Yang et~al.(2026{\natexlab{a}})Yang, Qin, Si, Chen, Gu, Yao, Lin, Wang, Wang, and Duan]{yang2026self}
Chenxu Yang, Chuanyu Qin, Qingyi Si, Minghui Chen, Naibin Gu, Dingyu Yao, Zheng Lin, Weiping Wang, Jiaqi Wang, and Nan Duan.
\newblock Self-distilled rlvr.
\newblock \emph{arXiv preprint arXiv:2604.03128}, 2026{\natexlab{a}}.

\bibitem[Yang et~al.(2026{\natexlab{b}})Yang, Liu, Xie, Yang, Yang, and Lin]{yang2026learning}
Wenkai Yang, Weijie Liu, Ruobing Xie, Kai Yang, Saiyong Yang, and Yankai Lin.
\newblock Learning beyond teacher: Generalized on-policy distillation with reward extrapolation.
\newblock \emph{arXiv preprint arXiv:2602.12125}, 2026{\natexlab{b}}.

\bibitem[Ye et~al.(2026{\natexlab{a}})Ye, Dong, Dong, Wu, Huang, and Wei]{ye2026online}
Tianzhu Ye, Li~Dong, Qingxiu Dong, Xun Wu, Shaohan Huang, and Furu Wei.
\newblock Online experiential learning for language models.
\newblock \emph{arXiv preprint arXiv:2603.16856}, 2026{\natexlab{a}}.

\bibitem[Ye et~al.(2026{\natexlab{b}})Ye, Dong, Wu, Huang, and Wei]{ye2026policy}
Tianzhu Ye, Li~Dong, Xun Wu, Shaohan Huang, and Furu Wei.
\newblock On-policy context distillation for language models.
\newblock \emph{arXiv preprint arXiv:2602.12275}, 2026{\natexlab{b}}.

\bibitem[Yu et~al.(2025)Yu, Zhang, Zhu, Yuan, Zuo, Yue, Dai, Fan, Liu, Liu, et~al.]{yu2025dapo}
Qiying Yu, Zheng Zhang, Ruofei Zhu, Yufeng Yuan, Xiaochen Zuo, Yu~Yue, Weinan Dai, Tiantian Fan, Gaohong Liu, Lingjun Liu, et~al.
\newblock Dapo: An open-source llm reinforcement learning system at scale.
\newblock \emph{arXiv preprint arXiv:2503.14476}, 2025.

\bibitem[Zeng et~al.(2026)Zeng, Lv, Hou, Du, Zheng, Chen, Yin, Ge, Xie, Wang, et~al.]{zeng2026glm}
Aohan Zeng, Xin Lv, Zhenyu Hou, Zhengxiao Du, Qinkai Zheng, Bin Chen, Da~Yin, Chendi Ge, Chengxing Xie, Cunxiang Wang, et~al.
\newblock Glm-5: from vibe coding to agentic engineering.
\newblock \emph{arXiv preprint arXiv:2602.15763}, 2026.

\bibitem[Zhao et~al.(2026{\natexlab{a}})Zhao, Xie, Wang, Li, Xie, and Sun]{zhao2026selfdistillationmultitokenprediction}
Guoliang Zhao, Ruobing Xie, An~Wang, Shuaipeng Li, Huaibing Xie, and Xingwu Sun.
\newblock Self-distillation for multi-token prediction, 2026{\natexlab{a}}.
\newblock URL \url{https://arxiv.org/abs/2603.23911}.

\bibitem[Zhao et~al.(2026{\natexlab{b}})Zhao, Xie, Liu, Huang, Pang, Chen, and Grover]{zhao2026self}
Siyan Zhao, Zhihui Xie, Mengchen Liu, Jing Huang, Guan Pang, Feiyu Chen, and Aditya Grover.
\newblock Self-distilled reasoner: On-policy self-distillation for large language models.
\newblock \emph{arXiv preprint arXiv:2601.18734}, 2026{\natexlab{b}}.

\bibitem[Zheng et~al.(2024)Zheng, Zhang, Zhang, Ye, Luo, Feng, and Ma]{zheng2024llamafactory}
Yaowei Zheng, Richong Zhang, Junhao Zhang, Yanhan Ye, Zheyan Luo, Zhangchi Feng, and Yongqiang Ma.
\newblock Llamafactory: Unified efficient fine-tuning of 100+ language models.
\newblock In \emph{Proceedings of the 62nd Annual Meeting of the Association for Computational Linguistics (Volume 3: System Demonstrations)}, Bangkok, Thailand, 2024. Association for Computational Linguistics.
\newblock URL \url{http://arxiv.org/abs/2403.13372}.

\end{thebibliography}
